\documentclass{article}

\PassOptionsToPackage{numbers, sort&compress}{natbib}

 \usepackage[main, final]{neurips_2026}

\usepackage[utf8]{inputenc}
\usepackage[T1]{fontenc}
\usepackage{hyperref}
\usepackage{url}
\usepackage{booktabs}
\usepackage{amsfonts}
\usepackage{nicefrac}
\usepackage{microtype}
\usepackage{xcolor}

\usepackage{graphicx}
\usepackage{wrapfig}
\usepackage{enumitem}

\usepackage{multirow}
\usepackage{amsmath, amssymb, amsthm, bm}
\usepackage{booktabs}
\usepackage{longtable}
\usepackage{array}
\usepackage{bm}
\usepackage{subcaption}
\usepackage{wrapstuff}
\usepackage{makecell}
\usepackage{placeins}
\usepackage{needspace}

\newtheorem{proposition}{Proposition}
\newtheorem{assumption}{Assumption}

\newcommand{\rev}[1]{#1}
\newcommand{\revo}[1]{#1}
\newcommand{\revcolor}{}
\InputIfFileExists{revmarks.tex}{}{}

\title{GeoWind2Plan: Mission-Time 3D Urban Wind Prediction for Energy-Efficient UAV Planning}
\hypersetup{pdftitle={GeoWind2Plan: Mission-Time 3D Urban Wind Prediction for Energy-Efficient UAV Planning},
  pdfauthor={Shaoxiang Qin, Yucheng Zhao, Fuyuan Lyu, Di Zhou, Jiachen Yao, Xue Liu, Anima Anandkumar, Liangzhu Leon Wang, Xiongye Xiao}}

\author{
  \parbox{\dimexpr\linewidth-2\tabcolsep-3pt\relax}{
  \centering
    Shaoxiang Qin\textsuperscript{1,2,3} \quad
    Yucheng Zhao\textsuperscript{1} \quad
    Fuyuan Lyu\textsuperscript{3} \quad
    Di Zhou\textsuperscript{1} \quad
    Jiachen Yao\textsuperscript{\revo{4}} \quad \\
    Xue Liu\textsuperscript{3,\revo{5}} \quad
    Anima Anandkumar\textsuperscript{\revo{4}}\footnotemark[1] \quad
    Liangzhu Leon Wang\textsuperscript{2}\footnotemark[1] \quad
    Xiongye Xiao\textsuperscript{1}\thanks{Corresponding authors: Anima Anandkumar (anima@caltech.edu); Liangzhu Leon Wang (leon.wang@concordia.ca); Xiongye Xiao (xxiao9@utk.edu)}
  }
  \\
  \parbox{\dimexpr\linewidth-2\tabcolsep-3pt\relax}{
  \centering
  \vspace{0.2 cm}
  \textsuperscript{1}University of Tennessee, Knoxville \quad
  \textsuperscript{2}Concordia University \quad
  \textsuperscript{3}McGill University \\
  \textsuperscript{\revo{4}}\revo{California Institute of Technology} \quad
  \textsuperscript{\revo{5}}\revo{MBZUAI}\\}
}

\begin{document}

\maketitle

\begin{abstract}
In urban low-altitude flight, buildings reshape ambient wind into spatially varying 3D flow, making unmanned aerial vehicle (UAV) energy depend on local wind exposure as well as path length. However, building-resolved wind information is rarely available when a mission must be planned. Computational fluid dynamics (CFD) can produce high-fidelity urban flow fields, but each simulation is tied to a fixed inflow boundary condition and can take hours to days, which is incompatible with urban UAV missions that typically last minutes to tens of minutes. We present \emph{GeoWind2Plan}, a geometry-to-wind-to-planning framework for mission-time 3D urban wind prediction and energy-efficient UAV planning. Given only a background wind vector, 3D building geometry, and a start-goal pair, GeoWind2Plan transforms the building geometry into a reference-wind frame, predicts mission-relevant 3D wind patches with a localized geometry-conditioned neural operator, stitches them into a queryable local wind field, and optimizes a feasible 3D path and speed profile using a physically grounded UAV energy model. Rather than pursuing CFD-perfect reconstruction, GeoWind2Plan targets decision-useful wind prediction: trajectories are planned with predicted wind and evaluated under high-fidelity CFD wind. Across held-out urban domains, wind speeds, and mission wind-angle regimes, GeoWind2Plan performs corridor-localized wind inference in about 3 seconds, compared with roughly 8 hours for CFD. Under CFD evaluation, trajectories planned with GeoWind2Plan reduce energy by 6.9\%, 12.7\%, and 4.5\% in tailwind, headwind, and crosswind missions relative to wind-agnostic planning, recovering 87.9\%, 85.7\%, and 75.0\% of CFD-reference savings. 
These results show that fast, corridor-localized 3D urban wind prediction can make wind-aware UAV energy planning practical at mission time. The code and dataset are available at \href{https://github.com/DUAL-Xiao/GeoWind2Plan/}{https://github.com/DUAL-Xiao/GeoWind2Plan/}. \rev{We provide an interactive demo at \href{https://geowind2plan.com}{https://geowind2plan.com}.}

\end{abstract}

\section{Introduction}
\label{sec:intro}
Low-altitude unmanned aerial vehicles (UAVs) are increasingly deployed for urban logistics, infrastructure inspection, mapping, environmental monitoring, and emergency response~\cite{yao2025unmanned,li2022application,smith2015use}. Across these applications, onboard energy remains a critical operational bottleneck that constrains flight range, payload capacity, and safety reserves~\cite{zhang2021energy,beigi2022overview}. Improving energy efficiency is therefore essential for extending mission capability, increasing robustness under uncertain operating conditions, and enabling scalable urban UAV operations. Wind makes this energy-planning problem fundamentally different from shortest-path planning: propulsion energy depends on the air-relative velocity along the trajectory, so a geometrically short route can be energy-inefficient if it exposes the vehicle to adverse wind conditions~\cite{rienecker2023energy}. In cities, buildings further reshape the ambient wind into spatially varying 3D flow: the same background wind can accelerate through street canyons, slow down in wakes, and turn around dense building clusters. As a result, a planner \revo{that} uses only a single wind vector or a height-only wind profile misses the local flow structures that determine the energy cost of a route. Prior work has shown that UAV trajectories in urban wind fields can exploit building-induced wind phenomena for lower energy consumption, but such methods typically assume that a realistic 3D wind field is already available to the planner~\cite{chan2023wind,ruijia2025integrating}. 

Computational fluid dynamics (CFD) can provide building-resolved 3D urban wind fields, but it is not a viable mission-time wind source~\citep{tominaga2008aij, shao2023pignn}. The difficulty is not only computational cost; it is also the fixed-boundary-condition nature of CFD. A CFD wind field is one solution for one city geometry under one inflow direction, speed, and boundary profile. When the background wind direction or speed changes, the boundary-value problem changes and a new simulation is required. Since wind direction is continuous and urban missions occur under changing weather conditions, precomputing or rerunning CFD for all possible wind conditions is infeasible. In our experiments, a high-resolution $1.2\,\mathrm{km}$ urban CFD case takes roughly 8 hours. This gap motivates a mission-time wind prediction framework that can adapt to a background inflow vector without rerunning CFD.

Learning-based flow prediction offers a promising route to this goal. Neural operators~\citep{kovachki2023neural,li2021fourier,li2024physics,gupta2021multiwavelet}, especially Fourier neural operators (FNOs)~\citep{li2021fourier}, learn mappings between function spaces and can approximate families of partial differential equation solutions much faster than classical solvers~\citep{berner2025principled}. Recent urban airflow models further show that localized training and building-aware geometric encodings, such as multi-directional distance features, can support high-resolution 3D wind prediction from limited CFD data~\cite{qin2025data}. However, turning a wind predictor into a mission-time UAV planning system still faces several key challenges: (i) high-fidelity 3D urban CFD labels are scarce, making it unrealistic to densely sample the product space of cities, wind directions, and wind speeds; (ii) high-resolution 3D wind fields contain millions of grid cells, while a UAV mission only needs the wind field in a task-relevant subset of the city; (iii) field-level accuracy alone does not determine planning value: a predicted wind field is useful only if it leads to a lower-energy trajectory when evaluated under the true urban wind.

\begin{figure}[t]
\centering
    \includegraphics[width=1\linewidth]{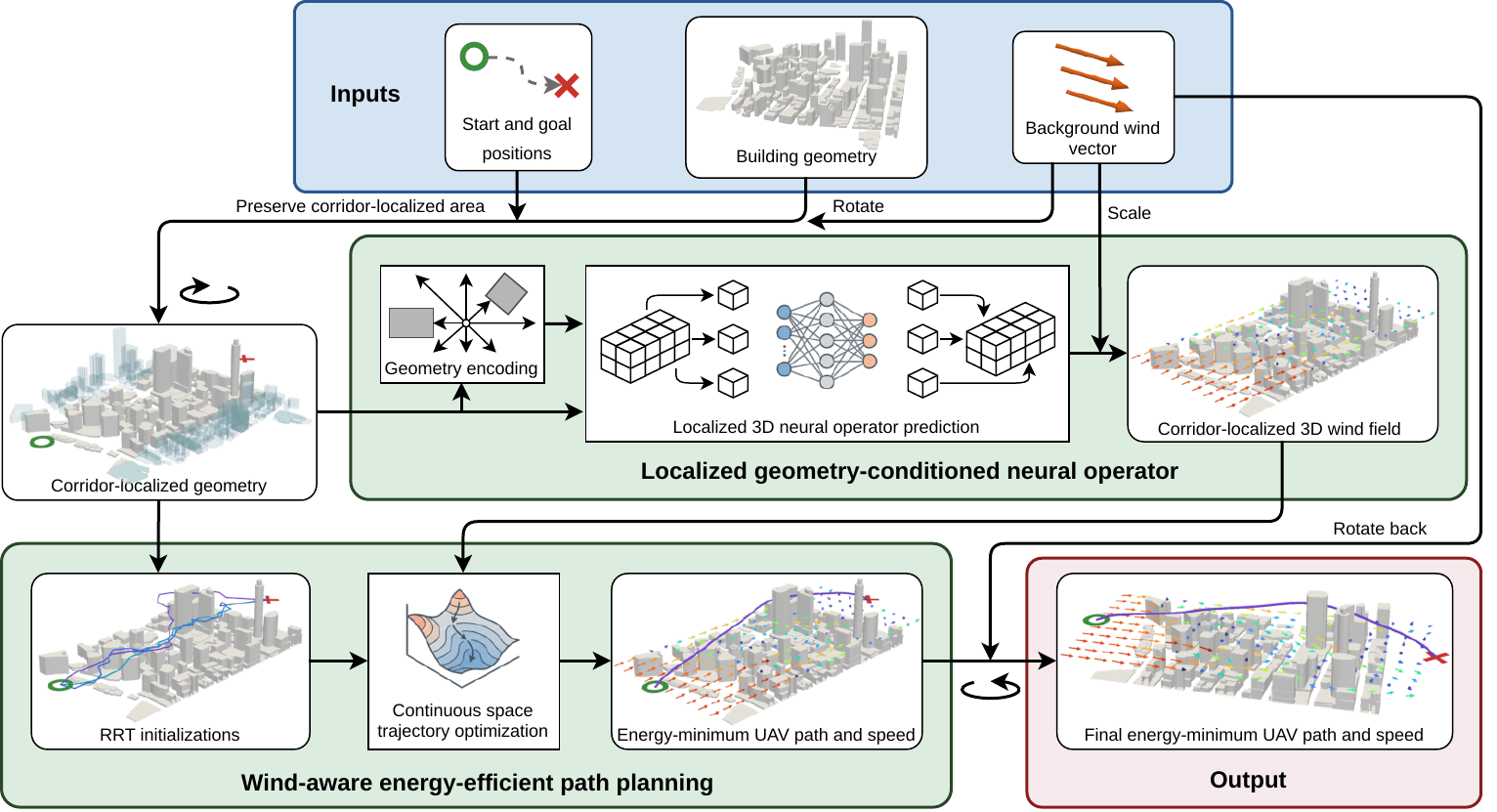}
\caption{\textbf{Overview of the GeoWind2Plan framework.} Given a start-goal pair, 3D building geometry, and a background wind vector, GeoWind2Plan extracts a mission-relevant corridor, rotates the local geometry into a reference-wind frame, predicts corridor-relevant wind patches with a localized geometry-conditioned neural operator, stitches them into a queryable local 3D wind field, rescales the field from the reference training speed to the given background wind speed, and rotates the predicted wind field back to the city frame. The resulting local wind field is used by an RRT-initialized continuous optimizer to produce an energy-efficient 3D UAV path and speed profile.}

\label{fig:framework}
\end{figure}

To address these challenges, we present \emph{GeoWind2Plan}, a mission-time geometry-to-wind-to-planning framework that couples 3D urban wind prediction with energy-efficient UAV trajectory optimization \rev{(Figure~\ref{fig:framework})}. GeoWind2Plan takes only a background wind vector, 3D building geometry, and a start-goal pair as input. To adapt to different incoming wind directions, it transforms the city geometry into a wind-aligned reference coordinate frame, applies a localized geometry-conditioned neural operator, and rotates the predicted vector field back to the original city frame. In terms of wind speed, GeoWind2Plan adapts to a mission background wind speed by multiplying the predicted mean wind field by the ratio between the background wind speed and the reference training speed. This rescaling is based on the mean-flow similarity of neutral high-Reynolds-number urban flows \citep{uehara2003critical,larose2006reynolds,chew2018flows}, where the mean flow normalized by the incoming wind speed is mainly controlled by building geometry and varies weakly with incoming wind speed. To make high-resolution 3D prediction practical at mission time, it evaluates only mission-relevant wind patches and stitches them into a queryable local wind field. This predicted field is then coupled with a physically grounded UAV energy model and \revo{a rapidly-exploring random tree (RRT)-initialized}~\cite{Lav98c} continuous trajectory optimizer to jointly optimize the 3D path and speed profile.

GeoWind2Plan is designed for decision-useful wind prediction rather than CFD-perfect reconstruction. We therefore evaluate the full geometry-to-wind-to-planning loop at the trajectory level: the planner optimizes trajectories using predicted wind, and the resulting trajectories are evaluated under high-fidelity CFD wind. Across held-out urban geometries, wind speeds, city-scale domains, and tailwind, headwind, and crosswind mission regimes, GeoWind2Plan performs corridor-localized wind inference in about 3 seconds compared with roughly 8 hours for CFD, while substantially reducing energy relative to both wind-agnostic planning and a height-only wind-profile baseline, and recovering most of the savings achieved by CFD-reference planning.

\paragraph{Contributions.}
To the best of our knowledge, GeoWind2Plan is the first mission-time learning-based framework that couples ambient-wind-conditioned, building-resolved 3D urban wind prediction with CFD-evaluated UAV energy planning. It provides three key capabilities:

\begin{enumerate}[leftmargin=1.35em, label=\arabic*., itemsep=2pt, topsep=2pt]
\item \textbf{Wind-conditioned geometric generalization.}
GeoWind2Plan learns a generalizable mapping from building geometry to urban wind
fields. The predictor is conditioned on 3D geometric features rather than on a
fixed city map, allowing it to infer local wind patches in unseen urban layouts.
Mission-specific wind directions are handled by rotating the city into a
reference-wind frame, so the model still performs geometry-conditioned inference
under a fixed reference inflow; the predicted vector field is then rotated back. This makes geometry, rather than a precomputed CFD field, the object that the model learns to generalize over.

\item \textbf{Mission-local 3D wind construction.}
GeoWind2Plan converts patch-level geometry-to-wind predictions into a
mission-specific 3D wind field. Given a start-goal pair, it evaluates only the
high-resolution patches intersecting the route-relevant corridor and stitches
them into a continuous, queryable wind field for planning. This avoids full-city
inference while preserving local urban flow structures such as wakes,
sheltering, channeling, and speed-up regions.

\item \textbf{Continuous wind-aware UAV trajectory planning.}
Unlike prior wind-aware planners that assume fixed speed or restrict routes to grid/graph edges, GeoWind2Plan plans the UAV trajectory and speed jointly in continuous 3D space. It uses multiple RRT warm starts to obtain feasible collision-free candidates, then refines each candidate by nonlinear trajectory optimization under wind-dependent energy and vehicle constraints. This allows the planner to adjust both position and speed to fully exploit local urban wind for lower energy.
\end{enumerate}

\section{Method}
\label{sec:methodology}

We present \emph{GeoWind2Plan}, a geometry-to-wind-to-planning framework for mission-time 3D urban wind prediction and energy-efficient UAV planning. We
start by formulating the coupled decision problem in
Section~\ref{sec:problem_setup}. Section~\ref{sec:wind_prediction} describes
the localized geometry-conditioned neural operator that converts background wind
and building geometry into a corridor-localized, queryable 3D wind field.
Section~\ref{sec:energy_model_main} defines the wind-aware multirotor energy
model used to evaluate candidate trajectories. Finally,
Section~\ref{sec:path_planning} presents the RRT-initialized continuous
optimizer that refines route, altitude, timing, and speed under vehicle and
obstacle constraints.

\subsection{Problem setup}
\label{sec:problem_setup}

We consider mission-time energy planning for a UAV operating in a 3D urban
domain \(\Omega \subset \mathbb{R}^3\). The city geometry is encoded by an
occupancy function \(\mathcal{G}\), where \(\mathcal{G}(\bm{x})=1\) denotes
building space and \(\mathcal{G}(\bm{x})=0\) denotes free space. A mission is
specified by a start location \(\bm{x}_s\) and a goal location \(\bm{x}_g\),
both in free space. Before planning, the only meteorological input is a
horizontal background wind vector
\(\bm{u}_{\mathrm{bg}} = U\bm{d}_{\theta}\), where
\(\bm{d}_{\theta}=(\cos\theta,\sin\theta,0)^\top\), \(U\) is the incoming wind
speed, and \(\theta\) is the incoming wind direction.

For a given city geometry and background wind, the true building-resolved urban
wind field is denoted by
\(\bm{w}_{\mathrm{CFD}}(\bm{x};\mathcal{G},\bm{u}_{\mathrm{bg}})\). This field
is used only for offline evaluation, because obtaining it with high-fidelity CFD
is too expensive for mission-time planning. During planning, the UAV does not
have access to \(\bm{w}_{\mathrm{CFD}}\). Instead, it must use the available
inputs \((\mathcal{G},\bm{u}_{\mathrm{bg}},\bm{x}_s,\bm{x}_g)\) to predict a
corridor-localized 3D wind field and then optimize a trajectory.

A trajectory is written as
\(\tau=\{\bm{x}(t),\bm{v}(t)\}_{t\in[0,\revo{t_{\mathrm f}}]}\), where
\(\bm{v}(t)=\dot{\bm{x}}(t)\) and \(\revo{t_{\mathrm f}}\) is the final mission time. It must connect \(\bm{x}_s\) to
\(\bm{x}_g\), avoid buildings with a prescribed clearance, and satisfy vehicle
limits on airspeed, acceleration, and thrust. We denote by \(E(\tau;\bm{w})\)
the physical energy of trajectory \(\tau\) under wind field \(\bm{w}\), and
write \(E_{\mathrm{CFD}}(\tau)=E(\tau;\bm{w}_{\mathrm{CFD}})\) for the final
CFD-evaluated energy.

The problem studied in this paper is a coupled geometry-to-wind-to-planning
decision problem:
\begin{equation}
    (\mathcal{G},\bm{u}_{\mathrm{bg}},\bm{x}_s,\bm{x}_g)
    \xrightarrow{\;\Pi_{\phi}\;}
    \widehat{\bm{w}}_{\mathrm{local}}
    \xrightarrow{\;\mathcal{P}\;}
    \tau^\star,
    \qquad
    \mathrm{score}(\tau^\star)
    =
    E_{\mathrm{CFD}}(\tau^\star).
    \label{eq:geometry_wind_planning_map}
\end{equation}
Here \(\Pi_{\phi}\) is the corridor-localized wind prediction module and
\(\mathcal{P}\) is the trajectory planner. The online planner computes
\begin{equation}
    \tau^\star
    \in
    \arg\min_{\tau\in\mathcal{T}(\mathcal{G},\bm{x}_s,\bm{x}_g)}
    E(\tau;\widehat{\bm{w}}_{\mathrm{local}}),
    \label{eq:planning_objective}
\end{equation}
where \(\mathcal{T}(\mathcal{G},\bm{x}_s,\bm{x}_g)\) denotes the set of
collision-free and dynamically feasible trajectories connecting the start and
goal. The final decision is evaluated by \(E_{\mathrm{CFD}}(\tau^\star)\), not
only by the pointwise error of \(\widehat{\bm{w}}_{\mathrm{local}}\). This
decision-level objective is central to our framework: the goal is not pointwise CFD reconstruction, but a fast, low-cost wind estimate that preserves the flow structures needed for energy-efficient planning at mission time.

\subsection{Localized geometry-conditioned neural operator for wind prediction}
\label{sec:wind_prediction}

\paragraph{Challenge.}
A CFD wind field is a solution for one city geometry under one inflow boundary
condition. When the background wind direction or speed changes, the physical
boundary-value problem changes and a new CFD simulation is required. Directly
learning over the product space of urban geometries, wind directions, and wind
speeds would require a prohibitive number of high-fidelity 3D CFD labels.
GeoWind2Plan addresses this challenge by separating the wind prediction module into
three parts: a reference-frame neural wind predictor, a physically motivated
wind-condition transformation, and corridor-localized patch inference.

\paragraph{Reference-frame neural wind predictor.}
Let \(\bm{e}_1=(1,0,0)^\top\) denote the reference inflow direction, and let
\(U_{\mathrm{ref}}\) be the reference inflow speed used in the training data.
The wind-prediction engine is a localized geometry-conditioned neural operator
\(F_{\phi}\), trained offline under the reference inflow
\(U_{\mathrm{ref}}\bm{e}_1\). Given building-aware geometric features in a local
3D patch, \(F_{\phi}\) predicts the time-averaged 3D velocity field in that
patch. In our implementation, \(F_{\phi}\) follows a localized FNO design with
multi-directional distance features, but in this paper it is used as one module
inside the full geometry-to-wind-to-planning framework rather than as an
isolated field surrogate.

\paragraph{Wind-direction adaptation by yaw covariance.}
The key structure used for wind-direction adaptation is the yaw covariance of
the building-resolved flow equations. Let \(Q_{\theta}\in \mathrm{SO}(3)\) be the yaw
rotation that maps the mission wind direction to the reference direction,
\(Q_{\theta}\bm{d}_{\theta}=\bm{e}_1\). For a scalar geometry field
\(\mathcal{G}\), define the rotated geometry
\begin{equation}
    (\mathcal{R}_{\theta}\mathcal{G})(\bm{y})
    =
    \mathcal{G}(Q_{\theta}^{\top}\bm{y}).
    \label{eq:rotated_geometry}
\end{equation}

\begin{proposition}[Yaw covariance of building-resolved mean flow]
\label{prop:yaw_covariance}
Let \(\mathcal{S}(\mathcal{G},U\bm{d})\) denote the unique steady or
ensemble-mean velocity solution of a neutral incompressible urban flow problem
with yaw-symmetric boundary conditions, no-slip building walls, and a
horizontally uniform inflow direction \(\bm{d}\). Then, for any yaw rotation
\(Q_{\theta}\) satisfying \(Q_{\theta}\bm{d}_{\theta}=\bm{e}_1\),
\begin{equation}
    \mathcal{S}(\mathcal{G},U\bm{d}_{\theta})(\bm{x})
    =
    Q_{\theta}^{\top}
    \mathcal{S}(\mathcal{R}_{\theta}\mathcal{G},U\bm{e}_1)
    (Q_{\theta}\bm{x}).
    \label{eq:yaw_covariance}
\end{equation}
\end{proposition}

Proposition~\ref{prop:yaw_covariance} is a covariance statement rather than an
invariance statement: rotating the city and inflow into a reference-wind frame
also rotates the vector wind field, so the predicted velocity vectors must be
rotated back to the original city coordinates. The proof follows from applying
a rigid yaw-coordinate transformation to the flow equations and boundary
conditions; it is provided in Appendix~\ref{app:yaw_covariance_proof}. In the learned framework, this covariance relation lets direction variation be represented as variation in the wind-aligned geometry seen by the reference-frame neural operator, rather than as a separate neural model for every inflow angle.

\paragraph{Wind-speed adaptation by mean-flow similarity.}
The reference-frame neural operator is trained at \(U_{\mathrm{ref}}\), while
the mission wind speed may be \(U\neq U_{\mathrm{ref}}\). We use the following
high-Reynolds-number mean-flow approximation.

\begin{assumption}[Reference-speed similarity of mean urban flow]
\label{assump:speed_similarity}
For the neutral, mechanically driven urban flows considered in this work, with
fixed city geometry, fixed incoming wind direction, and fixed inflow profile
shape, the normalized time-averaged mean wind field varies weakly with inflow
speed over the tested operating range. Thus,
\begin{equation}
    \mathcal{S}(\mathcal{G},U\bm{e}_1)(\bm{x})
    \approx
    \frac{U}{U_{\mathrm{ref}}}
    \mathcal{S}(\mathcal{G},U_{\mathrm{ref}}\bm{e}_1)(\bm{x}).
    \label{eq:speed_similarity}
\end{equation}
\end{assumption}

This assumption is not a claim that the Navier--Stokes equations behave linearly in
velocity. It only states that the normalized, time-averaged urban mean-flow
patterns relevant to the planner are approximately preserved across the
tested wind-speed range. Appendix~\ref{app:speed_generalization} provides the
physical basis and experimental validations of this approximation.

\paragraph{Building-aware geometry encoding.}
To provide the neural operator with geometric information, we convert the
occupancy field into a feature field
\(\mathcal{H}(\mathcal{G})\) consisting of occupancy, signed distance function
(SDF)~\cite{li2023geometry}, and multi-directional distance features (MDDFs)~\cite{qin2025data}: $\mathcal{H}(\mathcal{G})
    =
    \left[
    \mathcal{G},
    \mathrm{SDF}_{\mathcal{G}},
    \mathrm{MDDF}_{\mathcal{G}}^{1:K}
    \right]$. The SDF encodes distance to the nearest building surface, while the \revo{\(K\) MDDF channels} encode
distances to obstacles along multiple directions. These directional features
help the predictor represent local obstacle bearing, channeling, wake exposure,
and other geometry-induced structures that are difficult to infer from
occupancy alone.

\paragraph{Corridor-localized patch inference.}
High-resolution 3D urban wind fields contain millions of grid cells, whereas a
single UAV mission only requires wind information near feasible routes. Given
\((\bm{x}_s,\bm{x}_g)\), we define a task region
\(\Omega_{\mathrm{task}}(r)\) as a 3D corridor around the horizontal
\revo{start-goal} segment, extruded through the flight-altitude band. If \(d_{sg}\)
is the horizontal \revo{start-goal} distance, then the relative width \(r\) specifies
a total corridor width of \(r d_{sg}\).

After rotating the city into the reference-wind frame, GeoWind2Plan evaluates
only the prediction patches \rev{\(\Omega_j\)} that intersect the rotated task region,
\begin{equation}
    \mathcal{J}_{\mathrm{task}}(r)
    =
    \left\{
    j:
    \revo{\Omega_j} \cap Q_{\theta}\Omega_{\mathrm{task}}(r)
    \neq \emptyset
    \right\}.
    \label{eq:task_patch_set}
\end{equation}
The selected patch predictions are stitched, rotated back to the city frame,
and scaled by \(U/U_{\mathrm{ref}}\) to form a queryable local wind field
\(\widehat{\bm{w}}_{\mathrm{local}}\). The planner queries this field by
interpolation along continuous trajectories. In corridor-constrained
experiments, the same \(\Omega_{\mathrm{task}}(r)\) also restricts RRT
initialization and trajectory optimization, reducing mission-time inference
while preserving the wind structures available to the route search.

\subsection{UAV energy model in 3D wind fields}
\label{sec:energy_model_main}

\paragraph{Challenge.}
Energy evaluation for multirotor flight in urban wind must depend on air-relative motion, since local wind changes airspeed, drag, required thrust, and rotor power along a 3D trajectory. Existing models remain simplified: Ebert et~al.~\cite{ebert2023trajectory} use an empirical power--airspeed curve, while Gu et~al.~\cite{ruijia2025integrating} omit energy changes due to acceleration and altitude variation. 
These simplifications are insufficient for jointly optimizing 3D path geometry
and speed in spatially varying urban wind. We therefore derive a tractable
multirotor energy model from basic physical principles for variable-speed motion
in three-dimensional, time-averaged wind fields.

We discretize a trajectory into \(N\) nodes
\(\{(\bm{x}_i,\bm{v}_i,\bm{a}_i)\}_{i=0}^{N-1}\), where \(\bm{x}_i\),
\(\bm{v}_i\), and \(\bm{a}_i\) are position, ground-relative velocity, and
acceleration. For a wind field \(\bm{w}\), either the predicted wind during
planning or the CFD wind during evaluation, the air-relative velocity is $\bm{v}_{a,i}
    =
    \bm{v}_i-\bm{w}(\bm{x}_i)$. The parasite drag and required thrust are modeled as
\begin{equation}
    \bm{D}_i
    =
    -\frac{1}{2}\rho C_d A_f
    \|\bm{v}_{a,i}\|\bm{v}_{a,i},
    \qquad
    \bm{T}_i
    =
    m\bm{a}_i
    -
    m\bm{g}
    -
    \bm{D}_i .
    \label{eq:drag_thrust}
\end{equation}
Thus, the trajectory geometry, speed profile, acceleration, and local wind are
coupled through the required thrust. The nodal power is decomposed into useful mechanical power, induced power, and
blade profile power:
\begin{equation}
    P_i
    =
    \bm{T}_i^{\top}\bm{v}_{a,i}
    +
    P_{\mathrm{ind}}(\bm{T}_i,\bm{v}_{a,i})
    +
    P_{\mathrm{prof}}(\bm{T}_i).
    \label{eq:nodal_power}
\end{equation}
The rotor-loss terms follow standard momentum-theory approximations and are
given in Appendix~\ref{sec:energy_model}. The trajectory energy under wind
field \(\bm{w}\) is computed by trapezoidal integration:
\begin{equation}
    E(\tau;\bm{w})
    =
    \sum_{i=0}^{N-2}
    \frac{1}{2}
    \left(P_i+P_{i+1}\right)
    \Delta t_i .
    \label{eq:trajectory_energy}
\end{equation}

This model uses time-averaged wind fields and quasi-steady rotor power terms.
It does not explicitly model gusts, short-time turbulent fluctuations,
motor-controller transients, or vehicle-specific attitude dynamics. These
approximations keep the optimization tractable while capturing the main
mechanisms through which 3D urban wind affects UAV energy: air-relative speed,
drag, required thrust, altitude change, acceleration, and rotor losses.

\subsection{Energy-aware trajectory optimization}
\label{sec:path_planning}

\paragraph{Challenge.}
Existing wind-aware planners often restrict UAV motion to fixed altitude, fixed speed, or discrete graph nodes~\cite{rienecker2023energy, ebert2023trajectory, ruijia2025integrating}, which can produce unrealistic turns and inaccurate energy estimates. Pure continuous gradient-based optimization, however, can become trapped in local minima. 
GeoWind2Plan therefore combines sampling-based feasible
initialization with continuous energy optimization.

\paragraph{RRT warm starts.}
We first run RRT (rapidly-exploring random tree)~\cite{Lav98c} multiple times in the 3D free space to generate collision-free
candidate paths. The occupancy grid is used for obstacle checking, and the SDF
is used to enforce a clearance margin. These paths provide diverse feasible
initializations with different spatial layouts and wind exposure, but they are
not energy-optimized.

\paragraph{Continuous refinement.}
Each RRT path is resampled into \(N\) nodes and refined by nonlinear trajectory
optimization. For a planning wind field \(\bm{w}_p\), where
\(\bm{w}_p=\widehat{\bm{w}}_{\mathrm{local}}\) for GeoWind2Plan, the decision
variables are
\(\mathcal{Z}=\{\bm{x}_i,\bm{v}_i,\bm{a}_i\}_{i=0}^{N-1}\cup\{\Delta t\}\).
The optimizer solves
\begin{equation}
    \min_{\mathcal{Z}}
    \sum_{i=0}^{N-2}
    \frac{1}{2}
    \left(P_i+P_{i+1}\right)\Delta t,
    \qquad
    P_i
    =
    P\!\left(
    \bm{x}_i,\bm{v}_i,\bm{a}_i,\bm{w}_p(\bm{x}_i)
    \right).
    \label{eq:trajectory_optimization_objective}
\end{equation}

The constraints enforce boundary conditions, kinematic consistency, obstacle
clearance, and vehicle feasibility. Boundary constraints fix
\(\bm{x}_0=\bm{x}_s\) and \(\bm{x}_{N-1}=\bm{x}_g\). Trapezoidal collocation
enforces
\begin{equation}
    \bm{x}_{i+1}
    =
    \bm{x}_i
    +
    \frac{1}{2}
    \left(\bm{v}_i+\bm{v}_{i+1}\right)\Delta t,
    \qquad
    \bm{v}_{i+1}
    =
    \bm{v}_i
    +
    \frac{1}{2}
    \left(\bm{a}_i+\bm{a}_{i+1}\right)\Delta t .
    \label{eq:collocation}
\end{equation}
Collision avoidance is enforced by requiring the SDF at trajectory nodes and
selected intermediate segment samples to exceed a prescribed clearance. Vehicle
limits are enforced by
\begin{equation}
    \|\bm{v}_i-\bm{w}_p(\bm{x}_i)\|\le v_{\max},
    \qquad
    \|\bm{a}_i\|\le a_{\max},
    \qquad
    \|\bm{T}_i\|\le T_{\max}.
    \label{eq:vehicle_limits}
\end{equation}

The resulting nonlinear program is solved with an interior-point optimizer~\cite{wachter2006implementation} for
each RRT initialization, and the trajectory with the lowest planning energy is
selected. Because node positions are continuous variables, the final path can
smooth, shorten, climb, descend, or laterally shift relative to the RRT warm
start to exploit favorable wind structures while maintaining clearance and
dynamic feasibility. In all experiments, the selected trajectory is evaluated
under \(\bm{w}_{\mathrm{CFD}}\). Therefore, only the planning wind field varies
across methods; the energy model, constraints, optimizer, and CFD evaluation
protocol are held fixed.

\setlength{\textfloatsep}{11pt plus 2pt minus 2pt}
\section{Results}

\subsection{Experimental setup}
\label{sec:experimental_setup}

We evaluate GeoWind2Plan on four $1.2\,\mathrm{km}\times1.2\,\mathrm{km}$ standard urban blocks A--D and one $3\,\mathrm{km}\times3\,\mathrm{km}$ large urban block E. Appendix Figure~\ref{fig:inputgeometries}
shows the 3D building geometries of these evaluation domains. The default inflow is $4\,\mathrm{m/s}$. For wind-speed adaptation, urban block A is also tested at $2$ and $8\,\mathrm{m/s}$. \rev{For wind-direction adaptation, urban block A is tested with the incoming wind rotated by $30^\circ$ and $60^\circ$ from the reference direction.} Our model is trained offline on high-fidelity CFD wind fields; details are given in Appendix~\ref{sec:model_detail}. For each evaluation block, we sample 256 feasible start-goal missions near the horizontal domain boundaries and remove very short missions. The flight envelope is $30$--$120\,\mathrm{m}$, with both endpoints placed at $75\,\mathrm{m}$ by default. Let $\psi$ be the relative angle between the horizontal start-goal direction and the incoming wind. Missions are grouped as tailwind if ${|\psi|\le45^\circ}$, headwind if ${|\psi-180^\circ|\le45^\circ}$, and crosswind otherwise. 

To evaluate GeoWind2Plan, we use two mission-time baselines and one offline
reference. \textbf{Wind-agnostic} uses zero wind. \textbf{Vertical profile wind} uses a precomputed height-dependent wind profile obtained by averaging the urban-interior CFD training fields, and therefore captures vertical shear but no horizontal spatial structure. \revo{\textbf{Ours} uses the 3D wind field predicted by GeoWind2Plan's neural operator, which supports corridor-localized inference.} In the results, we use \textbf{Ground truth} to denote the offline
high-fidelity CFD wind field, which serves as the building-resolved reference
wind for evaluation and for an oracle planning setting; this label refers to the CFD reference wind input in our settings, not to a certified \revo{global} optimum. All four settings share the same missions, constraints, energy model, RRT warm starts, and continuous trajectory optimizer; only the planning wind input changes. Each method uses 10 RRT initializations per mission and selects the lowest-energy optimized trajectory under its own planning wind input. All evaluation blocks are held out and do not share CFD patches with the training set.

\subsection{Trajectory-level energy evaluation}
\label{sec:energy_performance}
\newcommand{\gain}[3]{#1\,\ensuremath{\pm}\,#2\,(#3\%\ensuremath{\uparrow})}
\newcommand{\plain}[2]{#1\,\ensuremath{\pm}\,#2}

\newcommand{\best}[1]{\textbf{#1}}

\newcommand{\tighttoprule}{\toprule}
\newcommand{\tightmidrule}{\midrule[\heavyrulewidth]}
\newcommand{\tightbottomrule}{\bottomrule}

\begin{table*}[!b]
\centering
\footnotesize
\setlength{\tabcolsep}{2.8pt}
\renewcommand{\arraystretch}{1.08}
\caption{Quantitative evaluation of UAV trajectory-planning efficiency across urban blocks A, B, and E at wind speed $U=4\,\mathrm{m/s}$ \revo{(block A also at $2$ and $8\,\mathrm{m/s}$)} and relative wind-angle regimes (see Appendix Table~\ref{tab:large_experiment_supp} for C and D). The table reports unit-distance energy consumption $(\mathrm{Wh}/\mathrm{km})$ as mean $\pm$ standard deviation, with extra energy measured relative to Ground truth reference. \rev{A ($30^\circ$) and A ($60^\circ$) denote urban block A with the incoming wind rotated by $30^\circ$ and $60^\circ$.} The best results are highlighted in bold. }
\label{tab:large_experiment}
\begin{tabular}{@{}cclllll@{}}
\tighttoprule
\multirow{2}{*}{{\bfseries\boldmath \makecell{Urban\\block}}} &
\multirow{2}{*}{{\bfseries\boldmath \revo{{\unboldmath$U$}}~$(\mathrm{m}/\mathrm{s})$}} &
\multirow{2}{*}{{\bfseries\boldmath Method}} &
\multicolumn{4}{c}{{\bfseries\boldmath Unit-distance energy consumption $(\mathrm{Wh}/\mathrm{km})$}} \\
\cmidrule{4-7}
& & & All & Tailwind & Headwind & Crosswind \\
\midrule

\multirow{12}{*}{A} & \multirow{4}{*}{2} & Ground truth   & \plain{12.4}{0.8} & \plain{11.3}{0.4} & \plain{13.2}{0.3} & \plain{12.5}{0.6} \\
\cmidrule{3-7}
 &  & Wind-agnostic & \gain{12.7}{0.8}{2.4} & \gain{11.7}{0.3}{3.1} & \gain{13.7}{0.4}{3.8} & \gain{12.7}{0.6}{1.5} \\
 &  & Profile wind   & \gain{12.6}{0.8}{1.3} & \gain{11.4}{0.4}{1.2} & \gain{13.4}{0.4}{2.1} & \gain{12.6}{0.6}{1.0} \\
 &  & \textbf{Ours} & \best{\gain{12.5}{0.8}{0.4}} & \best{\gain{11.3}{0.4}{0.2}} & \best{\gain{13.3}{0.3}{0.7}} & \best{\gain{12.5}{0.6}{0.3}} \\
\cmidrule{2-7}

 & \multirow{4}{*}{4} & Ground truth   & \plain{12.3}{1.4} & \plain{10.2}{0.6} & \plain{13.7}{0.5} & \plain{12.4}{1.0} \\
\cmidrule{3-7}
 &  & Wind-agnostic & \gain{13.3}{1.8}{7.5} & \gain{11.0}{0.4}{8.4} & \gain{15.4}{1.1}{12.1} & \gain{13.0}{1.2}{4.7} \\
 &  & Profile wind   & \gain{12.8}{1.6}{4.1} & \gain{10.6}{0.6}{3.8} & \gain{14.6}{0.7}{6.4} & \gain{12.8}{1.1}{3.0} \\
 &  & \textbf{Ours} & \best{\gain{12.5}{1.5}{1.4}} & \best{\gain{10.2}{0.5}{0.6}} & \best{\gain{14.0}{0.5}{2.1}} & \best{\gain{12.6}{1.0}{1.3}} \\
\cmidrule{2-7}

 & \multirow{4}{*}{8} & Ground truth   & \plain{12.5}{2.6} & \plain{8.5}{0.8} & \plain{15.2}{0.7} & \plain{12.6}{1.7} \\
\cmidrule{3-7}
 &  & Wind-agnostic & \gain{15.7}{4.7}{25.7} & \gain{10.3}{0.5}{20.9} & \gain{21.5}{3.2}{41.0} & \gain{14.9}{3.0}{17.9} \\
 &  & Profile wind   & \gain{14.1}{3.6}{13.0} & \gain{9.2}{0.8}{8.6} & \gain{18.1}{1.9}{18.9} & \gain{14.0}{2.2}{10.6} \\
 &  & \textbf{Ours} & \best{\gain{13.2}{3.0}{5.3}} & \best{\gain{8.6}{0.9}{1.7}} & \best{\gain{16.4}{1.0}{7.6}} & \best{\gain{13.2}{1.9}{4.8}} \\
\tightmidrule

\multirow{4}{*}{\rev{\makecell{A\\($30^\circ$)}}} & \multirow{4}{*}{\rev{4}} & \rev{Ground truth}   & \rev{\plain{12.4}{1.7}} & \rev{\plain{9.9}{0.7}} & \rev{\plain{14.1}{0.6}} & \rev{\plain{12.5}{1.1}} \\
\cmidrule{3-7}
 &  & \rev{Wind-agnostic} & \rev{\gain{13.6}{2.2}{9.0}} & \rev{\gain{10.8}{0.6}{9.6}} & \rev{\gain{16.3}{1.2}{14.9}} & \rev{\gain{13.2}{1.4}{5.7}} \\
 &  & \rev{Profile wind}   & \rev{\gain{13.1}{1.9}{5.0}} & \rev{\gain{10.4}{0.7}{5.2}} & \rev{\gain{15.1}{0.6}{6.9}} & \rev{\gain{13.0}{1.2}{3.9}} \\
 &  & \rev{\textbf{Ours}} & \rev{\best{\gain{12.6}{1.8}{1.5}}} & \rev{\best{\gain{10.0}{0.7}{1.1}}} & \rev{\best{\gain{14.4}{0.6}{2.1}}} & \rev{\best{\gain{12.7}{1.1}{1.3}}} \\
\tightmidrule

\multirow{4}{*}{\rev{\makecell{A\\($60^\circ$)}}} & \multirow{4}{*}{\rev{4}} & \rev{Ground truth}   & \rev{\plain{12.3}{1.5}} & \rev{\plain{10.3}{0.7}} & \rev{\plain{13.9}{0.5}} & \rev{\plain{12.3}{1.0}} \\
\cmidrule{3-7}
 &  & \rev{Wind-agnostic} & \rev{\gain{13.3}{2.0}{8.3}} & \rev{\gain{11.1}{0.6}{7.8}} & \rev{\gain{15.6}{1.2}{12.4}} & \rev{\gain{13.0}{1.2}{6.0}} \\
 &  & \rev{Profile wind}   & \rev{\gain{13.0}{1.8}{5.8}} & \rev{\gain{10.6}{0.7}{3.5}} & \rev{\gain{15.0}{0.9}{8.0}} & \rev{\gain{12.9}{1.2}{5.2}} \\
 &  & \rev{\textbf{Ours}} & \rev{\best{\gain{12.5}{1.6}{2.1}}} & \rev{\best{\gain{10.4}{0.7}{1.1}}} & \rev{\best{\gain{14.3}{0.5}{3.2}}} & \rev{\best{\gain{12.5}{1.0}{1.7}}} \\
\tightmidrule

\multirow{4}{*}{B} & \multirow{4}{*}{4} & Ground truth   & \plain{12.3}{1.7} & \plain{9.6}{0.5} & \plain{14.1}{0.7} & \plain{12.4}{1.1} \\
\cmidrule{3-7}
 &  & Wind-agnostic & \gain{13.6}{2.5}{10.3} & \gain{10.4}{0.5}{8.2} & \gain{16.6}{1.7}{18.0} & \gain{13.2}{1.5}{6.6} \\
 &  & Profile wind   & \gain{12.9}{2.0}{4.8} & \gain{10.1}{0.7}{4.4} & \gain{15.1}{0.8}{7.1} & \gain{12.9}{1.2}{3.6} \\
 &  & \textbf{Ours} & \best{\gain{12.6}{1.9}{1.9}} & \best{\gain{9.8}{0.6}{1.2}} & \best{\gain{14.5}{0.6}{2.7}} & \best{\gain{12.6}{1.2}{1.7}} \\
\tightmidrule

\multirow{4}{*}{E} & \multirow{4}{*}{4} & Ground truth   & \plain{12.2}{2.1} & \plain{8.9}{0.4} & \plain{14.3}{0.4} & \plain{12.4}{1.2} \\
\cmidrule{3-7}
 &  & Wind-agnostic & \gain{14.0}{3.2}{14.6} & \gain{9.8}{0.3}{10.5} & \gain{18.1}{0.9}{26.4} & \gain{13.5}{1.9}{8.9} \\
 &  & Profile wind   & \gain{12.7}{2.3}{3.8} & \gain{9.1}{0.4}{2.3} & \gain{15.2}{0.6}{6.3} & \gain{12.7}{1.4}{2.7} \\
 &  & \textbf{Ours} & \best{\gain{12.5}{2.3}{2.5}} & \best{\gain{9.0}{0.4}{0.7}} & \best{\gain{14.9}{0.5}{4.0}} & \best{\gain{12.6}{1.4}{2.2}} \\
\tightbottomrule

\end{tabular}
\end{table*}

Table~\ref{tab:large_experiment} reports CFD-evaluated unit-distance energy consumption in \revo{$\mathrm{Wh}/\mathrm{km}$}\rev{, defined as the trajectory energy divided by the straight-line start-goal distance}. Values are mean \(\pm\) standard deviation over missions, and percentages in parentheses denote energy overhead relative to \emph{Ground truth}. For a $1.2\,\mathrm{km}$ urban case, generating the CFD wind field takes roughly 8 hours, 
whereas our neural operator takes 15 seconds for full-field prediction and 3 seconds for 20\% corridor-width prediction (see Section~\ref{sec:corridor_localization}). Vertical profile wind is precomputed and has no online cost. For each planning instance, one RRT initialization followed by trajectory optimization took \revo{about} 2 seconds on a CPU core. Multiple RRT initializations are independent \revo{and can run in parallel}.

\begin{figure}[!t]
\centering
\includegraphics[width=0.98\columnwidth]{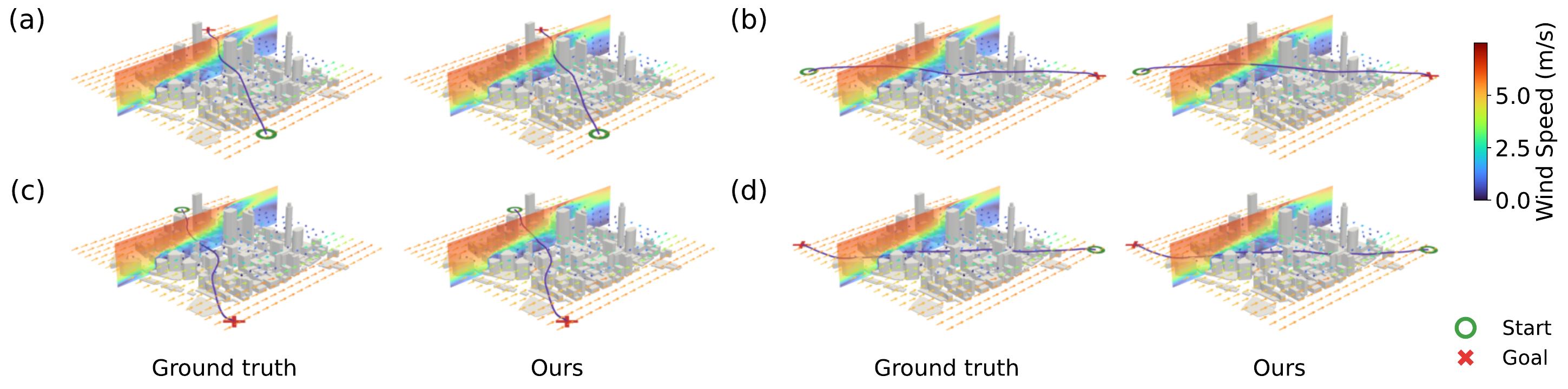}
\caption{Qualitative trajectory-planning comparison in urban block D. Panels (a)--(d) show four randomly selected \revo{start-goal} missions with different relative wind angles. Each pair compares the trajectory planned with the offline CFD reference
wind field (\textbf{Ground truth}) and the trajectory planned with GeoWind2Plan's predicted 3D wind field (\textbf{Ours}). \revo{A vertical wind-speed slice} and horizontal \revo{wind} vectors at \(75\,\mathrm{m}\) are shown as context, with optimized
UAV trajectories overlaid.}
\label{fig:3d_path}
\end{figure}

\textbf{Across urban blocks.} Table~\ref{tab:large_experiment} and Appendix Table~\ref{tab:large_experiment_supp} evaluate the urban blocks at the default wind speed. Across layouts, tailwind missions use the least energy, headwind missions use the most and show the largest method gap, and crosswind missions fall between. The method order is stable: GeoWind2Plan is closest to Ground truth, vertical profile wind gives a weaker improvement, and wind-agnostic planning is worst. Averaged over the four \revo{standard blocks A--D}, Ours reduces energy over wind-agnostic planning by $6.9\%$, $12.7\%$, and $4.5\%$ in tailwind, headwind, and crosswind missions, recovering $87.9\%$, $85.7\%$, and $75.0\%$ of the corresponding Ground truth reduction. This shows that the predicted 3D wind field preserves decision-useful flow structure across building layouts. Figure~\ref{fig:3d_path} and Appendix Figure~\ref{fig:additional_traj} compare trajectories planned with GeoWind2Plan and Ground truth reference.

\textbf{Across wind speeds.}
The wind-speed experiments on urban block A test whether mean-flow rescaling can reuse the reference-speed wind estimate at other inlet speeds. This rescaling is an approximate high-Reynolds-number similarity assumption. As speed increases from $2$ to $8\,\mathrm{m/s}$, UAV energies do not rise uniformly: stronger tailwinds can lower energy, while stronger headwinds raise it. Thus,{\parfillskip=0pt\par} \begin{wrapstuff}[r,width=0.60\textwidth,type=figure]
\centering
\includegraphics[width=\linewidth]{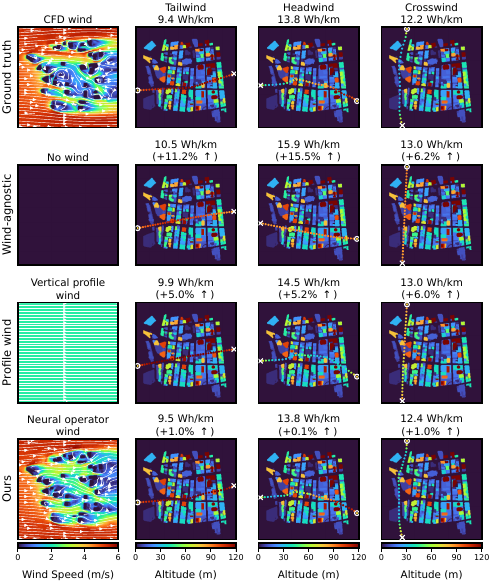}
\caption{Representative optimized trajectories (columns 2--4) under tailwind, headwind, and crosswind. Left column: horizontal wind field at start \revo{height} ($75\,\mathrm{m}$). White circle/cross: start/goal. Trajectory points and building background colored by altitude.}
\label{fig:path}
\end{wrapstuff}
Ground truth changes only mildly in the all-regime average, but inaccurate planning wind fields become much more costly. In the clearest case, the $8\,\mathrm{m/s}$ headwind regime, the increase over Ground truth is $41.0\%$ for wind-agnostic planning and $18.9\%$ for vertical profile wind, but only $7.6\%$ for Ours. Stronger wind therefore increases the value of local 3D wind structure. Appendix~\ref{app:validation_speed_rescaling} further supports mean-flow rescaling \revo{at the wind-field level}.

\leavevmode\rev{\textbf{Across wind directions.}
The neural operator is trained only with the incoming wind along the reference direction, and other directions are handled by the yaw rotation in Section~\ref{sec:wind_prediction}. For the two rotated directions on urban block A, Ground truth comes from separate CFD simulations, and Ours uses the same trained model. Without retraining, Ours stays close to Ground truth in all three directions. Averaged over the directions, the all-regime increase over Ground truth is $1.7\%$ for Ours, compared with $8.3\%$ for wind-agnostic planning and $5.0\%$ for vertical profile wind. The gap is largest in headwind missions. With the incoming wind rotated by $30^\circ$, the increase is $14.9\%$ for wind-agnostic planning and $6.9\%$ for vertical profile wind, but only $2.1\%$ for Ours.}

\textbf{Across urban scale.}
Urban block E tests the same pipeline at a larger scale. Without changing the model or planner, Ours stays close to Ground truth, with a $2.5\%$ all-regime increase versus $14.6\%$ for wind-agnostic planning. Vertical profile wind is also closer on this block, with a $3.8\%$ increase, likely because more open or weakly obstructed airspace makes the wind field less spatially complex and the height-only approximation less harmful. Ours still gives the closest result to Ground truth, \revo{showing that the 3D wind prediction of GeoWind2Plan remains useful at larger urban scale.}

\newsavebox{\tabtwobox}
\sbox{\tabtwobox}{\footnotesize\setlength{\tabcolsep}{3pt}%
\rev{\begin{tabular}{@{}lccc@{}}
\toprule
 & \multicolumn{3}{c}{{\bfseries\boldmath Energy $(\mathrm{Wh}/\mathrm{km})$}} \\
\cmidrule{2-4}
{\bfseries\boldmath Method} & Tailwind & Headwind & Crosswind \\
\midrule
Ground truth   & \plain{10.2}{0.6} & \plain{13.7}{0.5} & \plain{12.4}{1.0} \\
\cmidrule{1-4}
Wind-agnostic  & \plain{11.0}{0.4} & \plain{15.4}{1.1} & \plain{13.0}{1.2} \\
Profile wind   & \plain{10.6}{0.6} & \plain{14.6}{0.7} & \plain{12.8}{1.1} \\
Potential flow & \plain{10.6}{0.6} & \plain{14.6}{0.7} & \plain{12.8}{1.1} \\
Heuristic flow & \plain{10.5}{0.6} & \plain{14.5}{0.7} & \plain{13.0}{2.9} \\
Transolver++   & \plain{10.7}{1.8} & \plain{14.5}{0.6} & \plain{12.8}{1.1} \\
\textbf{Ours}  & \best{\plain{10.2}{0.5}} & \best{\plain{14.0}{0.5}} & \best{\plain{12.6}{1.0}} \\
\bottomrule
\end{tabular}}}
\begin{wraptable}[14]{r}{\wd\tabtwobox}
\vspace{-13.4pt}
\centering
\small
\caption{\rev{Comparison with other wind estimates on urban block A at $U=4\,\mathrm{m/s}$ (full results in Appendix Table~\ref{tab:additional_baselines}).}}
\label{tab:additional_baselines_main}
\usebox{\tabtwobox}
\end{wraptable}

\rev{\textbf{Other wind estimates.} We next test the energy performance of other wind estimates that use the same inputs. We consider two fast physics-based estimates and one learned baseline. The physics-based estimates are potential flow, which corrects vertical profile wind so that the flow passes around buildings but has no wakes, and heuristic flow, which imposes empirical wake and street-canyon velocity fields around each building~\cite{singh2008evaluation}. The learned baseline is Transolver++~\cite{DBLP:conf/icml/LuoWZXD0L25}, a neural solver trained on the same training blocks as our neural operator. Table~\ref{tab:additional_baselines_main} compares their energy consumption with that of Ours on urban block A. Potential flow performs like vertical profile wind, since its simplified physics cannot reproduce the complex flow structures around buildings. Heuristic flow uses slightly less energy than vertical profile wind in tailwind and headwind missions but more in crosswind missions. Its gains are inconsistent because the wakes estimated by empirical rules are not accurate. Transolver++ performs about the same as vertical profile wind, since it is not specifically designed for training efficiently on scarce data.\par}

\begin{figure}[!t]
\centering
\includegraphics[width=\linewidth]{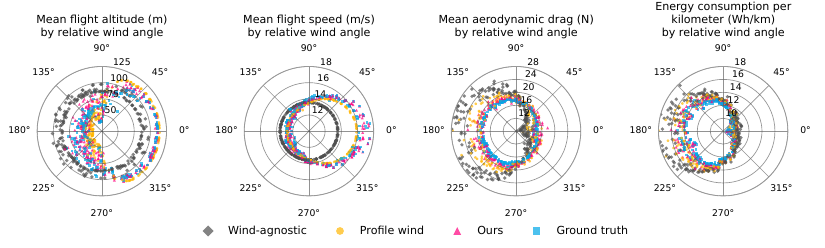}
\caption{Trajectory statistics by relative wind angle $\psi$, defined as the angle between the horizontal \revo{start-goal} direction and the incoming wind direction. Tailwind: ${|\psi|\le45^\circ}$; headwind: ${|\psi-180^\circ|\le45^\circ}$; crosswind: otherwise. Point angle denotes relative wind angle, while radius denotes the corresponding statistic\rev{, with flight speed measured relative to the ground}.}
\label{fig:trajectory_polar}
\end{figure}

\subsection{Trajectory adaptation behind energy savings}
\label{sec:wind_use_mechanisms}

Figure~\ref{fig:path} shows representative top-view trajectories under tailwind, \revo{headwind, and crosswind} missions, with path color indicating altitude. Wind-agnostic planning has no wind input and therefore stays close to the shortest feasible route. Vertical profile wind can use the fact that urban wind generally becomes stronger with height, so it climbs in tailwind missions and descends in headwind missions. However, because it has no horizontal spatial structure, its projected routes remain largely direct. Ours produces the subtle lateral bending seen in Ground truth, showing that the predicted 3D wind field provides the missing horizontal variation needed to choose locally lower-energy exposure.

Figure~\ref{fig:trajectory_polar} shows that the trajectory changes in Figure~\ref{fig:path} are systematic across relative wind angles. Wind-agnostic planning keeps nearly constant mean altitude and ground speed because its objective ignores wind direction. Wind-aware methods instead fly higher and faster in tailwinds, and lower and slower in headwinds. The drag plot shows that this is not simple drag minimization. Total energy integrates the full power model over time, and a multirotor continuously spends power to remain airborne. In tailwinds, Ours and Ground truth choose higher ground speed; mean drag can be higher than that of wind-agnostic planning, but shorter mission time lowers total energy. In headwinds, they choose lower speed; flight time increases slightly, but lower air-relative velocity and drag-related load reduce integrated energy. See Appendix Figure~\ref{fig:additional_traj_4paths} for more trajectory comparisons.

\wrapstuffclear
\subsection{Effect of corridor localization}
\label{sec:corridor_localization}

\newsavebox{\tabthreebox}
\sbox{\tabthreebox}{\footnotesize\setlength{\tabcolsep}{4pt}%
\begin{tabular}{@{}lcc@{}}
\toprule
{\bfseries\boldmath Method} & {\bfseries\boldmath Corridor width} & {\bfseries\boldmath Energy $(\mathrm{Wh}/\mathrm{km})$} \\
\midrule
Ground truth   & Full field & \plain{12.3}{1.4} \\
\cmidrule{1-3}
Wind-agnostic  & Full field & \plain{13.3}{1.8} \\
Ours           & Full field & \plain{12.5}{1.5} \\
Ours           & 50\%       & \plain{12.4}{1.5} \\
Ours           & 20\%       & \plain{12.4}{1.6} \\
Ours           & 10\%       & \plain{12.4}{1.9} \\
\bottomrule
\end{tabular}}
\begin{wraptable}{r}{\wd\tabthreebox}
\vspace{-13.4pt}
\centering
\caption{\revo{Unit-distance energy consumption for different corridor widths on urban block A, averaged over all relative wind-angle regimes.}}
\label{tab:corridor_energy}
\usebox{\tabthreebox}
\vspace{-10pt}
\end{wraptable}

Table~\ref{tab:corridor_energy} compares full-field planning with corridor widths of $r=50\%$, $20\%$, and $10\%$. 
Inference time scales roughly with $r$. A \(20\%\) corridor reduces inference from \(15\,\mathrm{s}\) to \(3\,\mathrm{s}\). CFD-evaluated energy changes little across these settings, suggesting that energy-optimal routes usually stay near the direct start-goal line. Thus, a narrow corridor can preserve the relevant wind information while reducing inference time. The main cost is feasibility: at $r=10\%$, failures increase when a building taller than $120\,\mathrm{m}$ spans the corridor and blocks all feasible paths.

\section{Conclusion}
\label{sec:conclusion}

GeoWind2Plan shows that energy-efficient urban UAV planning does not require pointwise-perfect wind reconstruction. It learns a generalizable geometry-to-wind mapping, predicts the mission-relevant 3D wind corridor, and couples this local wind field with continuous wind-aware trajectory optimization. Across diverse urban geometries, wind speeds, and wind-angle regimes, GeoWind2Plan produces trajectories close to CFD-reference planning while reducing wind inference from hours to seconds, turning building-resolved urban wind into a practical mission-time planning signal.

\begin{ack}

This work was supported in part by an AI Tennessee seed grant, which contributed to the development of the research reported in this manuscript. Shaoxiang Qin, Yucheng Zhao\rev{,} and Xiongye Xiao acknowledge support from the U.S. National Science Foundation (NSF) through the Science and Technology Center for Complex Particle Systems (COMPASS) [Award No. 2243104]. 
Anima Anandkumar is supported in part by Bren endowed chair, ONR (MURI grant N00014-23-1-2654), and the AI2050 senior fellow program at Schmidt Sciences. Liangzhu Leon Wang acknowledges financial support from the Natural Sciences and Engineering Research Council of Canada (NSERC) through the Discovery Grants Program [RGPIN-2024-06297] and the Canada First Research Excellence Fund (CFREF) [IMPACT Project – Transforming Built and Urban Microclimates: Advancing Resilience Science for Vulnerable Populations in a Decarbonized and Electrified Canada]\rev{, and computing support from the ``Digital Research Alliance of Canada and Calcul Québec'' clusters}. The authors declare no competing interests.
\end{ack}

\bibliographystyle{unsrtnat}
{\small
\bibliography{neurips_2026}
}

\newpage
\appendix
\makeatletter
\let\gwtp@addcontentsline\addcontentsline
\renewcommand{\addcontentsline}[3]{\def\gwtp@a{#1}\def\gwtp@b{toc}\ifx\gwtp@a\gwtp@b\gwtp@addcontentsline{atc}{#2}{#3}\else\gwtp@addcontentsline{#1}{#2}{#3}\fi}
\setcounter{tocdepth}{2}
\section*{Appendix contents}
{\hyphenpenalty=10000\exhyphenpenalty=10000%
\renewcommand*\l@section[2]{\ifnum\c@tocdepth>\z@\addpenalty\@secpenalty\addvspace{1.0em\@plus\p@}\setlength\@tempdima{1.5em}%
\begingroup\parindent\z@\rightskip\@pnumwidth\@plus 6em\parfillskip-\@pnumwidth\leavevmode\bfseries\advance\leftskip\@tempdima\hskip-\leftskip
#1\nobreak\hfil\nobreak\hb@xt@\@pnumwidth{\hss #2\kern-\p@\kern\p@}\par\endgroup\fi}%
\@starttoc{atc}}
\makeatother
\clearpage
\section{Limitations}
\label{app:limitations}
This work uses time-averaged velocity fields. The current energy model does not capture turbulence-induced power fluctuations, and the planner does not account for turbulence-related safety risks. Extending the predictor to turbulence quantities, such as turbulent kinetic energy, is a direct next step. We also assume that the background wind direction and speed remain fixed during a mission. Future work aims to support faster wind updates and replanning under time-varying inflow conditions.

\section{Broader impacts}
\label{app:broader_impacts}

This work may have positive impacts on urban UAV operations by improving energy efficiency, mission range, and operational robustness in building-resolved wind fields. By enabling mission-time 3D urban wind prediction and wind-aware trajectory planning, the proposed framework can reduce the need for expensive online CFD simulations and make energy-efficient UAV planning more practical for applications such as urban logistics, infrastructure inspection, environmental monitoring, mapping, and emergency response.

At the same time, more efficient urban UAV planning may also encourage larger-scale deployment of low-altitude drones in cities. Such deployment could increase noise, contribute to low-altitude airspace congestion, and raise privacy concerns. These risks should be considered when applying wind-aware UAV planning methods in real urban environments.

\section{Related work}

\subsection{Neural operators for PDE surrogate modeling}

Deep learning has become an important tool for accelerating PDE-based simulations. Physics-informed neural networks~\cite{raissi2019physics} incorporate governing-equation residuals and boundary conditions into the training objective, reducing the need for labeled simulation data, but they often face optimization difficulties and scalability issues on large computational domains~\cite{hao2024pinnacle,liu2024preconditioning}. In contrast, operator-learning methods aim to approximate the mapping from input functions to solution fields. DeepONet~\cite{lu2021learning} learns this mapping through branch and trunk networks, while \rev{the} \revo{Fourier neural operator} (FNO)~\cite{li2021fourier} performs the mapping efficiently in the Fourier domain. This field-to-field formulation makes neural operators more suitable than standard CNNs or MLPs for PDE surrogate modeling, where the output is a continuous spatial field rather than a fixed-dimensional prediction~\cite{kovachki2023neural,berner2025principled,duruisseaux2025fourier}.

Among neural operators, FNO~\cite{li2021fourier} is especially attractive for grid-based fluid problems because it captures global interactions through spectral-domain operations and has been applied to large-scale physical systems such as climate modeling~\cite{kurth2023fourcastnet} and turbulent flow prediction~\cite{li2021fourier}. PINO~\cite{li2024physics} adds PDE-based constraints to the operator-learning loss for physics-consistent training. Geo-FNO~\cite{li2023fourier} extends Fourier operators to non-rectangular domains through geometric transformations. GINO~\cite{li2023geometry} combines geometry encoding with graph and Fourier operators to handle large-scale 3D PDEs with varying geometries. In parallel, graph-based neural PDE solvers~\cite{brandstetter2022message,pfaff2020learning} represent discretized domains as graphs and use message passing to model PDE dynamics on irregular meshes.

\subsection{Deep learning for urban wind field prediction}

Deep learning urban airflow surrogate modeling uses urban geometry and boundary conditions to rapidly predict wind fields that would otherwise require expensive CFD simulations. Existing studies have used GANs~\cite{huang2022accelerated}, CNN/U-Net models~\cite{clemente2024rapid, wang2025evaluating}, MLP-Mixers~\cite{clarke2025deep}, physics-informed GNNs~\cite{shao2023pignn}, memory-scalable GNNs~\cite{liu2023accurate}, and localized Fourier neural operators~\cite{qin2025data} for urban wind prediction.

However, several practical limitations remain. First, many fast raster-based methods formulate the problem as fixed-size two-dimensional image-to-image prediction~\cite{huang2022accelerated,clemente2024rapid,wang2025evaluating,clarke2025deep}. This is efficient for pedestrian-level assessment, but it cannot directly recover full 3D flow structures, which are important for applications involving vertical ventilation, height-dependent wind distribution, and building wake effects.

Second, most existing models require large CFD-generated datasets~\cite{huang2022accelerated,clemente2024rapid,wang2025evaluating,clarke2025deep,shao2023pignn,liu2023accurate}. This limits practical deployment because high-quality CFD data are expensive to generate, especially for real urban geometries, multiple wind conditions, and large computational domains. A surrogate model that depends on thousands of simulations may therefore be difficult to adapt to new cities or project-specific design contexts.

Third, many studies train mainly on idealized, parameterized, or synthetically generated urban geometries~\cite{huang2022accelerated,clemente2024rapid,wang2025evaluating,clarke2025deep,shao2023pignn,liu2023accurate}. Such datasets are useful for controlled experiments, but they may not fully represent the irregular geometry, density variation, and long-range aerodynamic interactions found in real urban districts. This can reduce reliability when the model is applied to complex unseen sites.

Finally, scalability remains challenging. Fixed-size image models are naturally tied to a prescribed resolution and domain size~\cite{huang2022accelerated,clemente2024rapid,wang2025evaluating,clarke2025deep}, while mesh-based GNNs require careful memory management for large 3D urban domains~\cite{liu2023accurate}. This makes large-area wind prediction difficult when only limited GPU memory is available.

\subsection{Wind-aware UAV planning}

Energy-efficient UAV planning in urban wind fields has been studied, but existing methods usually simplify either wind prediction or rotary-wing energy modeling. Rienecker et~al.~\cite{rienecker2023energy} coupled PALM large-eddy simulation with an A*-based planner to exploit urban wind for energy saving. However, the method targets a fixed-wing UAV flying at constant true airspeed, and the wind field is produced by offline high-fidelity simulation. Folk et~al.~\cite{folk2025towards} studied real-time energy-aware planning for rotary-wing UAVs using MPPI, but the wind field was generated by a simplified two-dimensional flow solver, and the energy cost relied on an empirical power–airspeed relation.

For rotary-wing UAVs in CFD-based urban wind fields, Ebert et~al.~\cite{ebert2023trajectory} reconstructed RANS wind fields with Gappy POD and used them for \revo{energy-aware} planning. Yet the planning was limited to a two-dimensional wind slice, and the energy model was only an empirical power curve. Gu et~al.~\cite{ruijia2025integrating} proposed a more detailed urban logistics framework with CFD wind fields, wind-resistance and turbulence constraints, and a \revo{multirotor} power model including thrust, drag, induced velocity, payload, airspeed, and \revo{ground speed}. Still, the planning is mainly for two-dimensional steady level cruise, without explicit modeling of acceleration, vertical motion, or changes in kinetic and potential energy.

Learning-based wind prediction has also been used to reduce CFD cost. Veiga-Piñeiro et~al.~\cite{veiga2025hybrid} trained a CAE-DNN surrogate from RANS data to predict urban wind speed and turbulent kinetic energy for turbulence-aware A* planning. However, the planner focuses on risk avoidance rather than energy minimization, and it does not include rotary-wing energy modeling or joint speed–trajectory optimization.

\section{Additional \revo{experimental} results}

\subsection{Additional task trajectory visualization}

\begin{figure}[tp]
    \centering
    \includegraphics[
        width=\linewidth,
        height=1.0\textheight,
        keepaspectratio
    ]{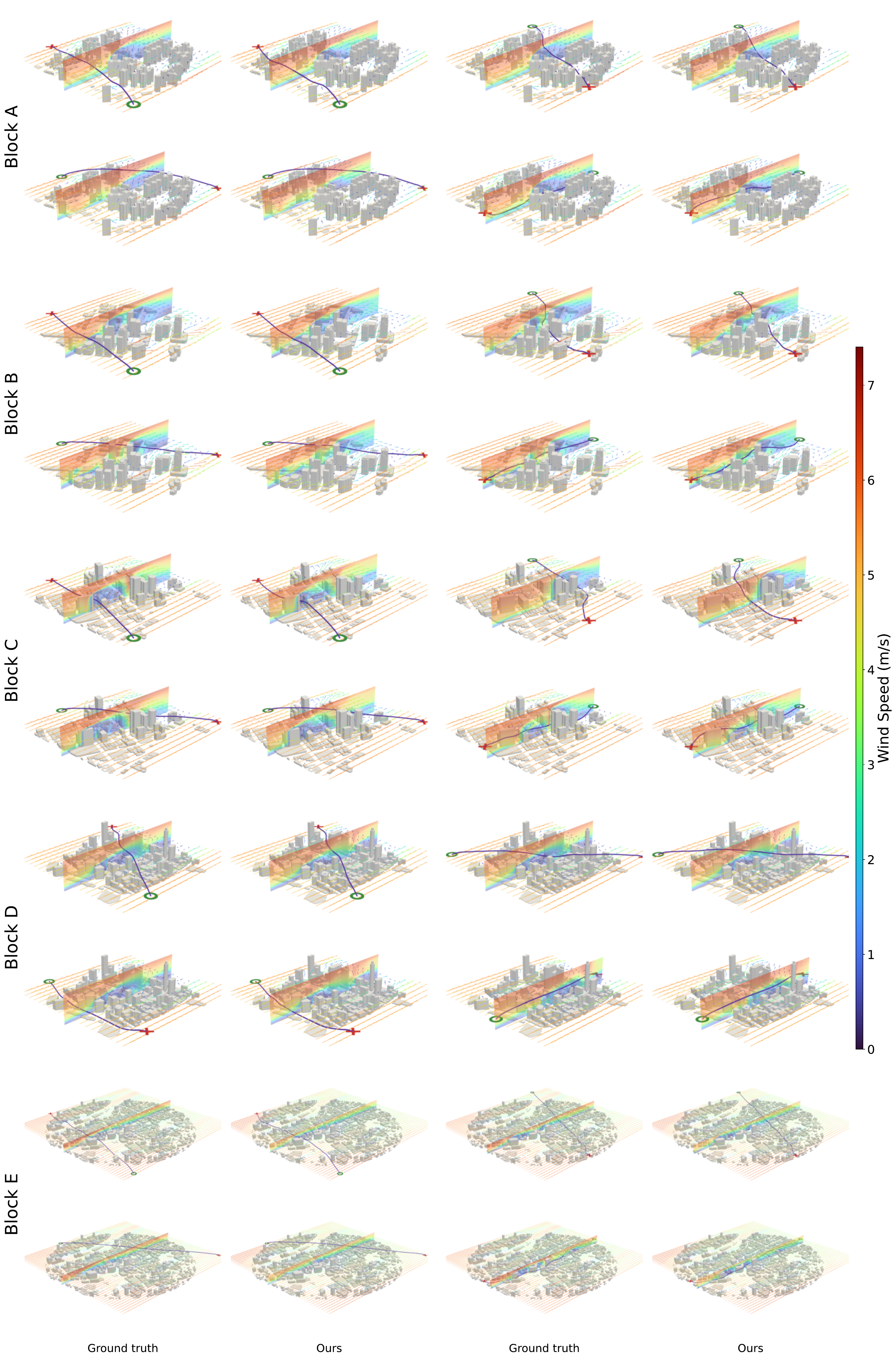}
    \caption{Additional task trajectory visualization over urban blocks A--E. Four tasks are randomly selected from each urban block, demonstrating generalizability across different urban environments and \revo{start-goal} settings.}
    \label{fig:additional_traj}
\end{figure}

Figure~\ref{fig:additional_traj} provides additional qualitative examples of task trajectories across urban blocks A--E. For each block, four \revo{start-goal} tasks are randomly selected to compare the ground-truth trajectories with the trajectories generated by our method. The results show that GeoWind2Plan produces paths that closely follow the reference solutions across diverse urban layouts and different \revo{start-goal} configurations, further demonstrating its generalization ability in unseen navigation scenarios.

\begin{figure}[tp]
    \centering
    \includegraphics[
        width=\linewidth,
        height=1.0\textheight,
        keepaspectratio
    ]{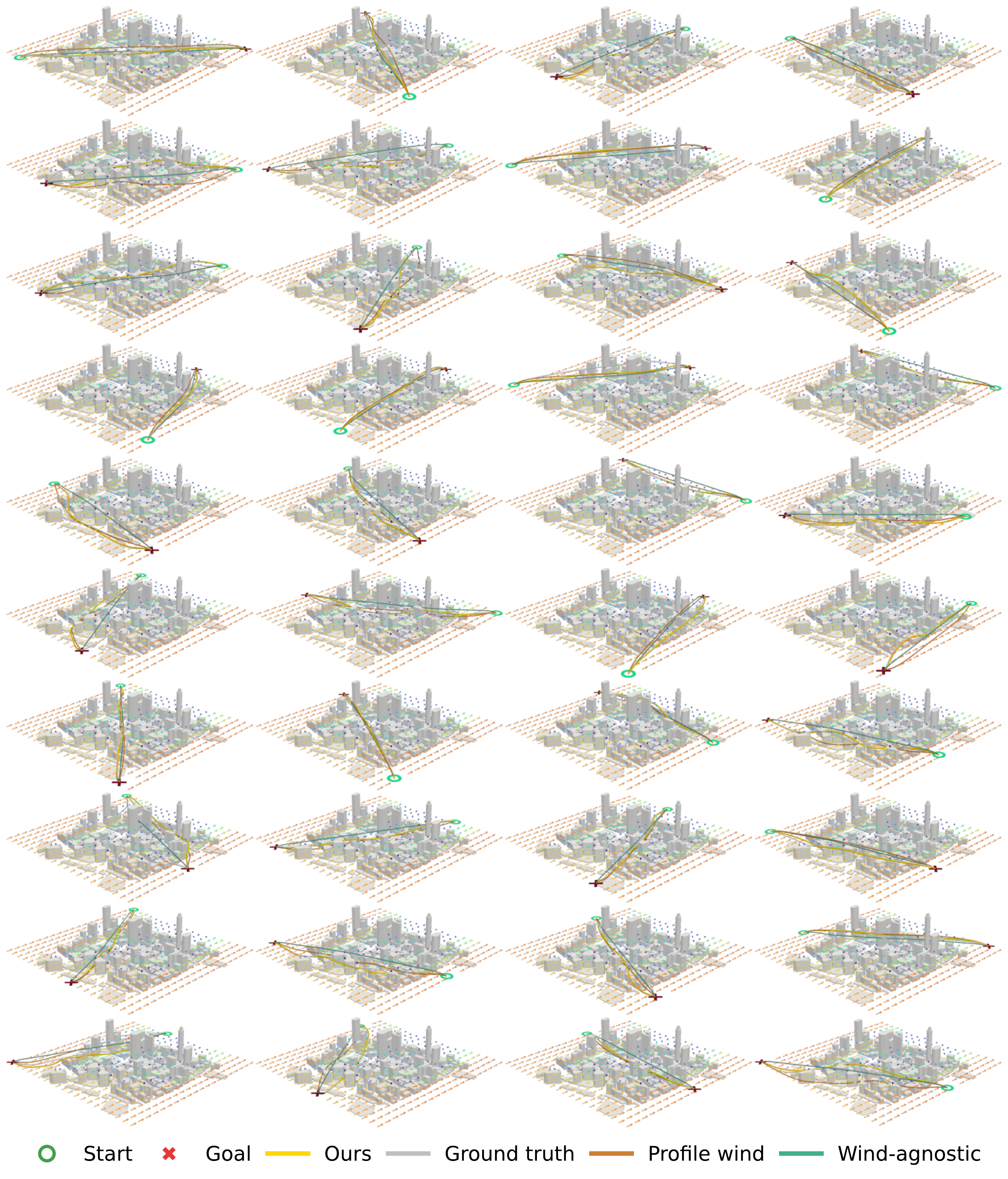}
    \caption{Additional trajectory-planning comparisons across 40 randomly sampled missions in urban block D. Each panel overlays trajectories planned with four different planning-time wind inputs: offline CFD reference wind (\textbf{\revo{Ground truth}}), GeoWind2Plan-predicted local 3D wind (\textbf{Ours}), height-only profile wind (\textbf{\revo{Profile wind}}), and zero wind (\textbf{\revo{Wind-agnostic}}).}
    \label{fig:additional_traj_4paths}
\end{figure}

Figure~\ref{fig:additional_traj_4paths} is an overview of diverse UAV flight trajectories generated across 40 randomly sampled scenarios in a complex urban environment (\revo{urban block D}). The figure shows that trajectories planned with GeoWind2Plan generally follow the CFD-reference routes more closely than the profile-wind and wind-agnostic baselines.

\clearpage

\subsection{Additional path planning results}

\begin{table*}[htp]
\centering
\footnotesize
\setlength{\tabcolsep}{2.8pt}
\renewcommand{\arraystretch}{1.08}
\caption{Quantitative evaluation of UAV trajectory-planning efficiency across five urban blocks (A--E) at wind speed $U=4\,\mathrm{m/s}$ \revo{(block A also at $2$ and $8\,\mathrm{m/s}$)} and relative wind-angle regimes. The table reports unit-distance energy consumption $(\mathrm{Wh}/\mathrm{km})$ as mean $\pm$ standard deviation, with extra energy measured relative to Ground truth reference. The best results are highlighted in bold.}
\label{tab:large_experiment_supp}
\begin{tabular}{@{}cclllll@{}}
\tighttoprule
\multirow{2}{*}{{\bfseries\boldmath \makecell{Urban\\block}}} &
\multirow{2}{*}{{\bfseries\boldmath \revo{{\unboldmath$U$}}~$(\mathrm{m}/\mathrm{s})$}} &
\multirow{2}{*}{{\bfseries\boldmath Method}} &
\multicolumn{4}{c}{{\bfseries\boldmath Unit-distance energy consumption $(\mathrm{Wh}/\mathrm{km})$}} \\
\cmidrule{4-7}
& & & All & Tailwind & Headwind & Crosswind \\
\midrule

\multirow{12}{*}{A} & \multirow{4}{*}{2} & Ground truth   & \plain{12.4}{0.8} & \plain{11.3}{0.4} & \plain{13.2}{0.3} & \plain{12.5}{0.6} \\
\cmidrule{3-7}
 &  & Wind-agnostic & \gain{12.7}{0.8}{2.4} & \gain{11.7}{0.3}{3.1} & \gain{13.7}{0.4}{3.8} & \gain{12.7}{0.6}{1.5} \\
 &  & Profile wind   & \gain{12.6}{0.8}{1.3} & \gain{11.4}{0.4}{1.2} & \gain{13.4}{0.4}{2.1} & \gain{12.6}{0.6}{1.0} \\
 &  & \textbf{Ours} & \best{\gain{12.5}{0.8}{0.4}} & \best{\gain{11.3}{0.4}{0.2}} & \best{\gain{13.3}{0.3}{0.7}} & \best{\gain{12.5}{0.6}{0.3}} \\
\cmidrule{2-7}

 & \multirow{4}{*}{4} & Ground truth   & \plain{12.3}{1.4} & \plain{10.2}{0.6} & \plain{13.7}{0.5} & \plain{12.4}{1.0} \\
\cmidrule{3-7}
 &  & Wind-agnostic & \gain{13.3}{1.8}{7.5} & \gain{11.0}{0.4}{8.4} & \gain{15.4}{1.1}{12.1} & \gain{13.0}{1.2}{4.7} \\
 &  & Profile wind   & \gain{12.8}{1.6}{4.1} & \gain{10.6}{0.6}{3.8} & \gain{14.6}{0.7}{6.4} & \gain{12.8}{1.1}{3.0} \\
 &  & \textbf{Ours} & \best{\gain{12.5}{1.5}{1.4}} & \best{\gain{10.2}{0.5}{0.6}} & \best{\gain{14.0}{0.5}{2.1}} & \best{\gain{12.6}{1.0}{1.3}} \\
\cmidrule{2-7}

 & \multirow{4}{*}{8} & Ground truth   & \plain{12.5}{2.6} & \plain{8.5}{0.8} & \plain{15.2}{0.7} & \plain{12.6}{1.7} \\
\cmidrule{3-7}
 &  & Wind-agnostic & \gain{15.7}{4.7}{25.7} & \gain{10.3}{0.5}{20.9} & \gain{21.5}{3.2}{41.0} & \gain{14.9}{3.0}{17.9} \\
 &  & Profile wind   & \gain{14.1}{3.6}{13.0} & \gain{9.2}{0.8}{8.6} & \gain{18.1}{1.9}{18.9} & \gain{14.0}{2.2}{10.6} \\
 &  & \textbf{Ours} & \best{\gain{13.2}{3.0}{5.3}} & \best{\gain{8.6}{0.9}{1.7}} & \best{\gain{16.4}{1.0}{7.6}} & \best{\gain{13.2}{1.9}{4.8}} \\
\tightmidrule

\multirow{4}{*}{B} & \multirow{4}{*}{4} & Ground truth   & \plain{12.3}{1.7} & \plain{9.6}{0.5} & \plain{14.1}{0.7} & \plain{12.4}{1.1} \\
\cmidrule{3-7}
 &  & Wind-agnostic & \gain{13.6}{2.5}{10.3} & \gain{10.4}{0.5}{8.2} & \gain{16.6}{1.7}{18.0} & \gain{13.2}{1.5}{6.6} \\
 &  & Profile wind   & \gain{12.9}{2.0}{4.8} & \gain{10.1}{0.7}{4.4} & \gain{15.1}{0.8}{7.1} & \gain{12.9}{1.2}{3.6} \\
 &  & \textbf{Ours} & \best{\gain{12.6}{1.9}{1.9}} & \best{\gain{9.8}{0.6}{1.2}} & \best{\gain{14.5}{0.6}{2.7}} & \best{\gain{12.6}{1.2}{1.7}} \\
\tightmidrule

\multirow{4}{*}{C} & \multirow{4}{*}{4} & Ground truth   & \plain{12.4}{2.0} & \plain{9.3}{0.4} & \plain{14.4}{0.8} & \plain{12.5}{1.3} \\
\cmidrule{3-7}
 &  & Wind-agnostic & \gain{13.9}{3.0}{12.1} & \gain{10.1}{0.3}{8.2} & \gain{17.5}{1.4}{21.6} & \gain{13.5}{1.9}{7.9} \\
 &  & Profile wind   & \gain{12.9}{2.3}{4.5} & \gain{9.6}{0.5}{3.0} & \gain{15.4}{0.8}{7.0} & \gain{12.9}{1.5}{3.5} \\
 &  & \textbf{Ours} & \best{\gain{12.6}{2.1}{1.6}} & \best{\gain{9.4}{0.4}{0.8}} & \best{\gain{14.7}{0.8}{2.6}} & \best{\gain{12.7}{1.4}{1.3}} \\
\tightmidrule

\multirow{4}{*}{D} & \multirow{4}{*}{4} & Ground truth   & \plain{12.2}{1.7} & \plain{9.6}{0.5} & \plain{13.9}{0.6} & \plain{12.4}{1.2} \\
\cmidrule{3-7}
 &  & Wind-agnostic & \gain{13.5}{2.4}{10.5} & \gain{10.5}{0.4}{9.2} & \gain{16.4}{1.3}{17.9} & \gain{13.2}{1.6}{6.9} \\
 &  & Profile wind   & \gain{12.8}{2.0}{4.7} & \gain{9.9}{0.5}{3.0} & \gain{14.9}{1.0}{7.1} & \gain{12.9}{1.4}{3.9} \\
 &  & \textbf{Ours} & \best{\gain{12.5}{1.9}{1.9}} & \best{\gain{9.7}{0.6}{0.8}} & \best{\gain{14.3}{0.8}{2.9}} & \best{\gain{12.6}{1.2}{1.7}} \\
\tightmidrule

\multirow{4}{*}{E} & \multirow{4}{*}{4} & Ground truth   & \plain{12.2}{2.1} & \plain{8.9}{0.4} & \plain{14.3}{0.4} & \plain{12.4}{1.2} \\
\cmidrule{3-7}
 &  & Wind-agnostic & \gain{14.0}{3.2}{14.6} & \gain{9.8}{0.3}{10.5} & \gain{18.1}{0.9}{26.4} & \gain{13.5}{1.9}{8.9} \\
 &  & Profile wind   & \gain{12.7}{2.3}{3.8} & \gain{9.1}{0.4}{2.3} & \gain{15.2}{0.6}{6.3} & \gain{12.7}{1.4}{2.7} \\
 &  & \textbf{Ours} & \best{\gain{12.5}{2.3}{2.5}} & \best{\gain{9.0}{0.4}{0.7}} & \best{\gain{14.9}{0.5}{4.0}} & \best{\gain{12.6}{1.4}{2.2}} \\
\tightbottomrule

\end{tabular}
\end{table*}

\begin{table*}[htp]
\centering
\caption {Additional statistics of flight characteristics under different wind conditions, \revo{start and goal} heights, and wind speeds, including altitude and speed.}
\label{tab:flight_stats_all_settings}
\footnotesize
\setlength{\tabcolsep}{3pt}
\renewcommand{\arraystretch}{1.08}
\begin{tabular}{lcclcccccccc}
\toprule
& & & 
& \multicolumn{4}{c}{{\bfseries\boldmath \revo{Mean flight altitude (m)}}} 
& \multicolumn{4}{c}{{\bfseries\boldmath \revo{Mean flight speed (m/s)}}} \\
\cmidrule(lr){5-8} \cmidrule(lr){9-12}
{\bfseries\boldmath \makecell{Urban\\block}} 
& {\bfseries\boldmath \makecell{Height\\(m)}} 
& {\bfseries\boldmath \makecell{\revo{{\unboldmath$U$}}\\\revo{(m/s)}}} 
& {\bfseries\boldmath Condition} 
& \makecell{Ground\\truth} & \makecell{\revo{Wind-}\\\revo{agnostic}} & \makecell{Profile\\wind} & Ours 
& \makecell{Ground\\truth} & \makecell{\revo{Wind-}\\\revo{agnostic}} & \makecell{Profile\\wind} & Ours \\
\midrule
\multirow{20}{*}{A} 
& \multirow{4}{*}{50} 
& \multirow{4}{*}{4} 
& \revo{All}       & 71.4 & 66.8 & 63.1 & 71.4 & 13.91 & 13.53 & 13.82 & 13.92 \\
& & & Tailwind   & 100.7 & 65.5 & 108.2 & 102.3 & 15.71 & 13.53 & 15.50 & 15.89 \\
& & & Headwind   & 53.6 & 69.1 & 34.7 & 53.5 & 12.90 & 13.54 & 12.88 & 12.80 \\
& & & Crosswind  & 69.4 & 66.1 & 60.6 & 69.0 & 13.75 & 13.53 & 13.67 & 13.74 \\
\cmidrule{2-12}
& \multirow{12}{*}{75} 
& \multirow{4}{*}{2} 
& \revo{All}       & 86.7 & 87.6 & 75.0 & 86.1 & 13.65 & 13.54 & 13.66 & 13.68 \\
& & & Tailwind   & 110.5 & 87.7 & 113.1 & 110.3 & 14.57 & 13.52 & 14.55 & 14.68 \\
& & & Headwind   & 66.9 & 87.9 & 44.6 & 66.6 & 13.10 & 13.56 & 13.12 & 13.08 \\
& & & Crosswind  & 87.8 & 87.4 & 76.0 & 86.8 & 13.59 & 13.54 & 13.60 & 13.61 \\
\cmidrule{3-12}
& & \multirow{4}{*}{4} 
& \revo{All}       & 82.3 & 87.6 & 70.4 & 81.7 & 13.89 & 13.54 & 13.81 & 13.93 \\
& & & Tailwind   & 111.7 & 87.7 & 114.0 & 110.7 & 15.74 & 13.52 & 15.57 & 15.95 \\
& & & Headwind   & 65.4 & 87.9 & 39.7 & 62.8 & 12.86 & 13.56 & 12.81 & 12.78 \\
& & & Crosswind  & 79.9 & 87.4 & 69.5 & 80.4 & 13.72 & 13.54 & 13.66 & 13.75 \\
\cmidrule{3-12}
& & \multirow{4}{*}{8} 
& \revo{All}       & 77.0 & 87.6 & 64.7 & 75.9 & 14.45 & 13.54 & 14.10 & 14.49 \\
& & & Tailwind   & 112.0 & 87.7 & 114.2 & 109.6 & 18.11 & 13.52 & 17.64 & 18.47 \\
& & & Headwind   & 58.3 & 87.9 & 36.6 & 59.5 & 12.39 & 13.56 & 12.24 & 12.39 \\
& & & Crosswind  & 73.3 & 87.4 & 60.4 & 71.8 & 14.12 & 13.54 & 13.73 & 14.07 \\
\cmidrule{2-12}
& \multirow{4}{*}{100} 
& \multirow{4}{*}{4} 
& \revo{All}       & 91.6 & 111.9 & 77.8 & 89.4 & 13.88 & 13.54 & 13.79 & 13.92 \\
& & & Tailwind   & 117.2 & 111.5 & 118.5 & 116.4 & 15.75 & 13.54 & 15.64 & 16.03 \\
& & & Headwind   & 74.8 & 112.5 & 46.0 & 70.5 & 12.82 & 13.54 & 12.72 & 12.76 \\
& & & Crosswind  & 90.5 & 111.7 & 78.6 & 88.8 & 13.71 & 13.54 & 13.65 & 13.72 \\
\midrule[\heavyrulewidth]
\multirow{4}{*}{B} 
& \multirow{4}{*}{75} 
& \multirow{4}{*}{4} 
& \revo{All}       & 73.2 & 86.3 & 70.4 & 74.7 & 13.99 & 13.54 & 13.82 & 14.03 \\
& & & Tailwind   & 108.7 & 86.5 & 113.9 & 105.7 & 16.35 & 13.54 & 15.63 & 16.40 \\
& & & Headwind   & 54.3 & 86.2 & 40.7 & 57.7 & 12.64 & 13.54 & 12.80 & 12.71 \\
& & & Crosswind  & 69.6 & 86.2 & 69.1 & 71.8 & 13.80 & 13.54 & 13.66 & 13.81 \\
\midrule[\heavyrulewidth]
\multirow{4}{*}{C} 
& \multirow{4}{*}{75} 
& \multirow{4}{*}{4} 
& \revo{All}       & 70.2 & 87.8 & 70.6 & 72.1 & 13.99 & 13.55 & 13.82 & 14.03 \\
& & & Tailwind   & 110.4 & 87.3 & 114.1 & 109.5 & 16.59 & 13.55 & 15.64 & 16.70 \\
& & & Headwind   & 48.7 & 87.4 & 40.5 & 53.8 & 12.48 & 13.55 & 12.79 & 12.54 \\
& & & Crosswind  & 66.1 & 88.2 & 69.5 & 67.5 & 13.78 & 13.55 & 13.66 & 13.80 \\
\midrule[\heavyrulewidth]
\multirow{4}{*}{D} 
& \multirow{4}{*}{75} 
& \multirow{4}{*}{4} 
& \revo{All}       & 78.7 & 89.9 & 74.5 & 80.1 & 13.99 & 13.55 & 13.79 & 14.01 \\
& & & Tailwind   & 110.0 & 89.3 & 114.2 & 110.3 & 16.26 & 13.55 & 15.63 & 16.43 \\
& & & Headwind   & 60.8 & 91.0 & 46.1 & 59.6 & 12.70 & 13.55 & 12.74 & 12.67 \\
& & & Crosswind  & 76.1 & 89.6 & 74.0 & 79.1 & 13.79 & 13.55 & 13.63 & 13.78 \\
\midrule[\heavyrulewidth]
\multirow{4}{*}{E} 
& \multirow{4}{*}{75} 
& \multirow{4}{*}{4} 
& \revo{All}       & 68.3 & 89.9 & 69.6 & 70.7 & 13.95 & 13.44 & 13.76 & 13.91 \\
& & & Tailwind   & 115.9 & 90.2 & 117.5 & 113.6 & 16.90 & 13.44 & 15.57 & 16.88 \\
& & & Headwind   & 42.8 & 91.4 & 38.4 & 45.4 & 12.32 & 13.45 & 12.77 & 12.28 \\
& & & Crosswind  & 63.4 & 89.0 & 67.5 & 67.5 & 13.68 & 13.44 & 13.59 & 13.61 \\
\bottomrule
\end{tabular}
\end{table*}

\begin{figure}[tp]
    \centering
    \includegraphics[width=1.0\linewidth]{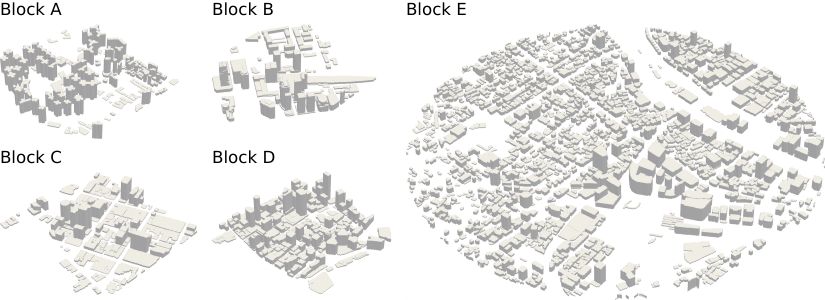}
    \caption{3D building geometry visualizations of the five randomly selected evaluation urban blocks used in the experiments: four standard \(1.2\,\mathrm{km}\times1.2\,\mathrm{km}\) blocks (A--D) and one larger \(3\,\mathrm{km}\times3\,\mathrm{km}\) block (E).}
    \label{fig:inputgeometries}
\end{figure}

\begin{table*}[tp]
\revcolor
\centering
\footnotesize
\setlength{\tabcolsep}{3pt}
\renewcommand{\arraystretch}{1.08}
\caption{\rev{Unit-distance energy consumption $(\mathrm{Wh}/\mathrm{km})$ with other wind estimates on urban blocks A and B at $U=4\,\mathrm{m/s}$. Values are mean $\pm$ standard deviation over missions, with extra energy measured relative to Ground truth reference. The best results are highlighted in bold.}}
\label{tab:additional_baselines}
\begin{tabular}{@{}clllll@{}}
\tighttoprule
\multirow{2}{*}{{\bfseries\boldmath \makecell{Urban\\block}}} &
\multirow{2}{*}{{\bfseries\boldmath Method}} &
\multicolumn{4}{c}{{\bfseries\boldmath Unit-distance energy consumption $(\mathrm{Wh}/\mathrm{km})$}} \\
\cmidrule{3-6}
& & All & Tailwind & Headwind & Crosswind \\
\midrule
\multirow{7}{*}{A} & Ground truth   & \plain{12.3}{1.4} & \plain{10.2}{0.6} & \plain{13.7}{0.5} & \plain{12.4}{1.0} \\
\cmidrule{2-6}
 & Wind-agnostic  & \gain{13.3}{1.8}{7.5} & \gain{11.0}{0.4}{8.4} & \gain{15.4}{1.1}{12.1} & \gain{13.0}{1.2}{4.7} \\
 & Profile wind   & \gain{12.8}{1.6}{4.1} & \gain{10.6}{0.6}{3.8} & \gain{14.6}{0.7}{6.4} & \gain{12.8}{1.1}{3.0} \\
 & Potential flow & \gain{12.8}{1.6}{4.1} & \gain{10.6}{0.6}{3.8} & \gain{14.6}{0.7}{6.4} & \gain{12.8}{1.1}{2.9} \\
 & Heuristic flow & \gain{12.9}{2.6}{4.6} & \gain{10.5}{0.6}{3.0} & \gain{14.5}{0.7}{5.6} & \gain{13.0}{2.9}{4.6} \\
 & Transolver++   & \gain{12.8}{1.7}{3.8} & \gain{10.7}{1.8}{4.9} & \gain{14.5}{0.6}{5.6} & \gain{12.8}{1.1}{2.6} \\
 & \textbf{Ours}  & \best{\gain{12.5}{1.5}{1.4}} & \best{\gain{10.2}{0.5}{0.6}} & \best{\gain{14.0}{0.5}{2.1}} & \best{\gain{12.6}{1.0}{1.3}} \\
\tightmidrule
\multirow{7}{*}{B} & Ground truth   & \plain{12.3}{1.7} & \plain{9.6}{0.5} & \plain{14.1}{0.7} & \plain{12.4}{1.1} \\
\cmidrule{2-6}
 & Wind-agnostic  & \gain{13.6}{2.5}{10.3} & \gain{10.4}{0.5}{8.2} & \gain{16.6}{1.7}{18.0} & \gain{13.2}{1.5}{6.6} \\
 & Profile wind   & \gain{12.9}{2.0}{4.8} & \gain{10.1}{0.7}{4.4} & \gain{15.1}{0.8}{7.1} & \gain{12.9}{1.2}{3.6} \\
 & Potential flow & \gain{12.9}{2.0}{4.7} & \gain{10.1}{0.7}{4.4} & \gain{15.1}{0.8}{7.0} & \gain{12.8}{1.2}{3.5} \\
 & Heuristic flow & \gain{12.9}{2.0}{4.6} & \gain{10.3}{1.7}{7.1} & \gain{15.0}{0.9}{6.3} & \gain{12.8}{1.2}{3.1} \\
 & Transolver++   & \gain{12.8}{2.0}{3.8} & \gain{9.9}{0.7}{3.0}  & \gain{15.0}{0.8}{6.1} & \gain{12.8}{1.2}{2.8} \\
 & \textbf{Ours}  & \best{\gain{12.6}{1.9}{1.9}} & \best{\gain{9.8}{0.6}{1.2}} & \best{\gain{14.5}{0.6}{2.7}} & \best{\gain{12.6}{1.2}{1.7}} \\
\tightbottomrule
\end{tabular}
\end{table*}

\subsection{Visualization of evaluation urban geometries}

We evaluate GeoWind2Plan on five randomly selected urban blocks: four standard
\(1.2\,\mathrm{km}\times1.2\,\mathrm{km}\) blocks (A--D) and one larger
\(3\,\mathrm{km}\times3\,\mathrm{km}\) block (E). Figure~\ref{fig:inputgeometries}
shows the 3D building geometries of these evaluation domains.

\rev{
\subsection{Comparison with other wind estimates}
\label{app:additional_baselines}

We compare GeoWind2Plan with three other wind estimates that, like GeoWind2Plan, use only the building geometry and the background wind. Each estimate replaces the planning wind field in the same planning and CFD-evaluation pipeline as in Table~\ref{tab:large_experiment}. The missions, energy model, constraints, and optimizer are unchanged.

\paragraph{Potential flow.} Potential flow is \revo{based on} the classical inviscid, irrotational estimate of flow around obstacles. Starting from the same height profile as vertical profile wind, it adds an irrotational correction that makes the flow mass-conserving and stops it from passing through buildings and the ground. The resulting flow is deflected around buildings but has no wakes or recirculation. The field is computed on a GPU in about $6\,\mathrm{s}$.

\paragraph{Heuristic flow.} Fast urban-wind models such as QUIC-URB build the flow from empirical rules. We implement the Modified R\"ockle and QUIC-URB parameterizations~\cite{singh2008evaluation}. Buildings are first decomposed into stacked rectangular blocks. Starting from the same height profile, the model imposes empirical velocity fields for upwind displacement, the leeward recirculation cavity and far wake, street-canyon vortices, sidewall recirculation, and a rooftop layer around each building. A final projection makes the field mass-conserving. The complete field is obtained in about $8\,\mathrm{s}$.

\paragraph{Transolver++.} We train Transolver++~\cite{DBLP:conf/icml/LuoWZXD0L25}, a state-of-the-art neural solver for partial differential equations on large 3D geometries, on the same training blocks as our neural operator. It predicts the complete time-averaged 3D wind field of an entire urban block at every grid point. To fit a complete $1.2\,\mathrm{km}$ block of 14.4 million grid points on a single GPU, we use a compact configuration with 4 layers, a hidden width of 40, 2 attention heads, and 8 slices.

Table~\ref{tab:additional_baselines} reports the full results on urban blocks A and B. The trends in Section~\ref{sec:energy_performance} hold on both blocks. On each block, potential flow performs almost the same as vertical profile wind in every regime, and neither heuristic flow nor Transolver++ consistently improves on it. Ours has the lowest energy in every regime on both blocks.

}

\section{Localized geometry-conditioned neural operator}
\label{sec:model_detail}

This section provides the implementation details of the localized geometry-conditioned neural operator used in GeoWind2Plan. Its purpose is to learn a mapping from building-resolved geometric features to local velocity fields under a reference inflow condition. The geometry encoding and model design are inspired by previous work in using 3D FNO for large-scale CFD simulations~\citep{li2023geometry, qin2025data}.

\paragraph{Model design.}
A high-resolution urban CFD field contains millions to tens of millions of grid cells. Learning a full-domain geometry-to-wind map directly would make each training sample extremely high-dimensional, require more CFD cases, and exceed the memory budget of modest GPUs. We therefore use a shared local neural operator on fixed-size 3D patches. Before cropping, the full urban geometry is encoded at every grid point by geometric features, including occupancy, SDF, \revo{and} MDDF, \revo{together with} coordinates, so each patch carries precomputed information about its surrounding building layout. The same local operator is then applied to all patches. It lifts the input features to a latent field, applies Fourier layers for spatial mixing, uses MLP blocks for channel mixing, and projects the output to the three wind-velocity components. At inference time, overlapping patch predictions are blended into a continuous wind field. This design reduces memory cost, turns each CFD case into many training samples, and allows prediction to be limited to task-relevant regions.

\paragraph{Geometry encoding.}
The input geometry is represented by occupancy, signed distance function (SDF)~\citep{li2023geometry}, \revo{and} multi-directional distance feature (MDDF)~\citep{qin2025data}, \revo{together with} spatial coordinates. The SDF gives the nearest signed distance to building surfaces. The MDDF complements this scalar distance by measuring, at each grid point, how far unobstructed space extends along a set of rays before reaching a building. We compute MDDF on the full domain before patch extraction, so obstacles outside a later cropped patch can still influence the feature value inside the patch. Two groups of directional rays are used: horizontal rays encode plan-view blockage and opening patterns, while wind-aligned vertical rays encode height-dependent obstruction along the main flow direction. To reduce channel count, the angular MDDF samples are transformed along the direction dimension, and only low-frequency components are retained. In our implementation, the ray directions are densely and uniformly sampled with several dozen angles, and \revo{no more than four} complete low-frequency angular modes are retained after compression. \rev{Each ray distance is capped at a length longer than the $1.2\,\mathrm{km}$ side of a standard block. In practice, almost no finite building distances are truncated, and the cap is reached primarily when a ray encounters no building and therefore replaces an otherwise infinite value.}

\paragraph{CFD data generation.}
Training labels are generated with CityFFD~\cite{mortezazadeh2022cityffd}, a GPU-based urban airflow solver that uses fast fluid dynamics with large-eddy simulation to resolve building-induced mean flow. CityFFD advances the flow with a semi-Lagrangian scheme and a pressure projection step, and uses high-order interpolation to reduce numerical diffusion on urban-scale grids. \rev{CityFFD has been validated against measured urban wind data~\cite{mortezazadeh2022estimating}.} We use it only to generate offline CFD labels; the planner does not call CityFFD at mission time. \rev{Each simulation covers a region much larger than the urban block. The block itself is resolved by a uniform $4\,\mathrm{m}\times4\,\mathrm{m}\times1.5\,\mathrm{m}$ grid, and it is surrounded by a buffer zone whose grid cells become progressively larger toward the edges of the simulated region. Following standard urban CFD practice, the buffer keeps the boundaries of the simulated region far from the buildings. Only the uniform-grid block is retained as data. Buildings are mapped onto this grid conservatively. Every cell that intersects a building is solid, and gaps narrower than one cell are closed.} The inlet wind follows a power-law profile with exponent $\alpha=0.15$ and reference height $z_{\mathrm{ref}}=10\,\mathrm{m}$\rev{, with a speed of $4\,\mathrm{m/s}$ at the reference height ($2$ and $8\,\mathrm{m/s}$ in the wind-speed tests on block A). Ground and building surfaces are no-slip, the lateral and top boundaries are symmetry boundaries, and the outlet has a constant pressure. Each simulation runs for 20{,}000 time steps of $0.5\,\mathrm{s}$, corresponding to a Courant number of about 0.5 at $4\,\mathrm{m/s}$. The first 10{,}000 steps are discarded, and the supervised target is the average of 500 velocity snapshots saved every 20 steps over the remaining steps.}

\rev{\paragraph{Dataset.} The dataset contains 33 urban blocks from real cities and is publicly available at \url{https://huggingface.co/datasets/UMBE-LAB/CityFFD-UrbanMicroclimate}. Table~\ref{tab:dataset} lists the blocks. The neural operator used in this paper is trained on 23 blocks, uses 3 further blocks only for early stopping, and does not see the remaining 7 blocks, including the evaluation blocks A--E, during training. The training, validation, and unseen blocks do not overlap and share no building geometry or CFD data, and all blocks are simulated with the same solver and boundary-condition setup. The wind-speed and wind-direction evaluations on block A use separate CFD simulations at $2$ and $8\,\mathrm{m/s}$ and with the incoming wind rotated by $30^\circ$ and $60^\circ$.}

\begin{table}[tp]
\revcolor
\centering
\small
\captionsetup{skip=7pt}
\caption{\rev{Urban blocks in the dataset. Training blocks are used to train the neural operator, validation blocks are used only for early stopping, and unseen blocks are not used in training. Letters A--E mark the evaluation blocks. Building coverage is the fraction of the block area covered by buildings, and height statistics are computed over all horizontal grid cells covered by buildings. Blocks marked with $^\dagger$ are $3\,\mathrm{km}\times3\,\mathrm{km}$, and all others are $1.2\,\mathrm{km}\times1.2\,\mathrm{km}$.}}
\label{tab:dataset}
\renewcommand{\arraystretch}{1.08}
\begin{tabular}{@{}lllcccc@{}}
\toprule
\multirow{2}{*}{{\bfseries\boldmath Block}} & \multirow{2}{*}{{\bfseries\boldmath City}} & \multirow{2}{*}{{\bfseries\boldmath Use}} & \multirow{2}{*}{{\bfseries\boldmath \makecell{Building\\coverage}}} & \multicolumn{3}{c}{{\bfseries\boldmath Building height (m)}} \\
\cmidrule{5-7}
 & & & & Mean & Std. & Max \\
\midrule
bcn01 & Barcelona & Training & 0.33 & 18.8 & 19.2 & 114.0 \\
hkg01 & Hong Kong & Training & 0.29 & 37.9 & 37.8 & 193.5 \\
hkg02 & Hong Kong & Validation & 0.30 & 35.4 & 34.2 & 124.5 \\
hkg03 & Hong Kong & Training & 0.21 & 33.3 & 37.5 & 109.5 \\
hkg04 & Hong Kong & Training & 0.19 & 30.3 & 37.4 & 147.0 \\
hkg05 & Hong Kong & Unseen & 0.24 & 38.9 & 46.4 & 217.5 \\
hkg06 & Hong Kong & Training & 0.21 & 37.3 & 36.2 & 118.5 \\
hkg07 & Hong Kong & Training & 0.26 & 41.8 & 43.1 & 211.5 \\
hkg08 & Hong Kong & Unseen & 0.26 & 42.7 & 34.4 & 147.0 \\
hkg09 & Hong Kong & Training & 0.19 & 41.1 & 41.5 & 181.5 \\
hkg10 & Hong Kong & Training & 0.28 & 35.3 & 34.2 & 240.0 \\
hkg11 & Hong Kong & Training & 0.26 & 33.6 & 30.3 & 178.5 \\
hkg12 & Hong Kong & Training & 0.18 & 52.3 & 51.7 & 214.5 \\
hkg13 & Hong Kong & Unseen (A) & 0.18 & 52.4 & 54.1 & 144.0 \\
lon01 & London & Training & 0.45 & 38.6 & 50.8 & 240.0 \\
lon02 & London & Validation & 0.28 & 69.6 & 66.7 & 235.5 \\
mel01 & Melbourne & Training & 0.48 & 34.7 & 41.6 & 219.0 \\
mel02 & Melbourne & Unseen (C) & 0.36 & 22.5 & 35.9 & 193.5 \\
mel03 & Melbourne & Training & 0.43 & 44.1 & 40.3 & 240.0 \\
mel04 & Melbourne & Training & 0.27 & 44.8 & 31.9 & 142.5 \\
mtr01 & Montreal & Unseen (B) & 0.23 & 35.9 & 44.9 & 202.5 \\
nbo01 & Nairobi & Training & 0.17 & 14.1 & 27.2 & 184.5 \\
nbo02 & Nairobi & Training & 0.26 & 15.0 & 20.0 & 121.5 \\
osa01$^\dagger$ & Osaka & Unseen (E) & 0.27 & 33.0 & 26.0 & 169.5 \\
sfo01 & San Francisco & Training & 0.39 & 37.9 & 26.0 & 160.5 \\
syd01 & Sydney & Training & 0.33 & 39.9 & 41.8 & 240.0 \\
syd02 & Sydney & Unseen (D) & 0.34 & 39.6 & 42.1 & 240.0 \\
syd03 & Sydney & Training & 0.37 & 13.9 & 14.2 & 145.5 \\
syd04 & Sydney & Validation & 0.40 & 53.8 & 51.3 & 240.0 \\
syd05 & Sydney & Training & 0.28 & 19.3 & 24.3 & 157.5 \\
tor01 & Toronto & Training & 0.30 & 34.4 & 36.1 & 187.5 \\
tor02 & Toronto & Training & 0.37 & 33.0 & 34.0 & 162.0 \\
tyo01$^\dagger$ & Tokyo & Training & 0.27 & 33.7 & 29.7 & 180.0 \\
\bottomrule
\end{tabular}
\end{table}

\begin{figure}[tp]
    \centering
    \includegraphics[width=0.803\linewidth]{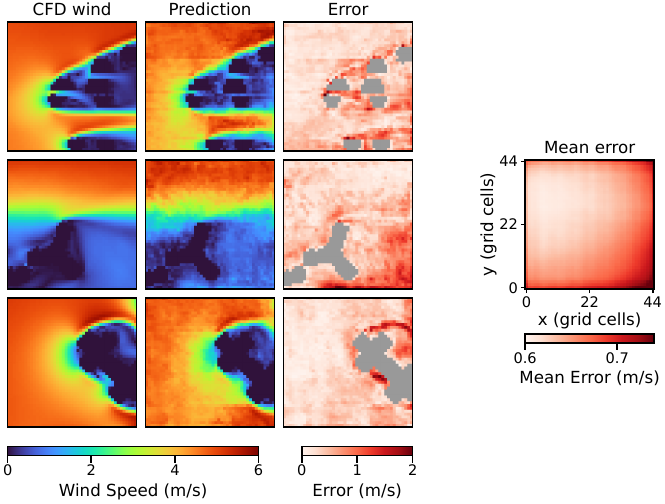}
    \caption{\rev{Patch predictions in urban block A without stitching. Left: three patches, each predicted individually. Each row shows the horizontal wind speed from CFD and from the patch prediction, and the magnitude of the velocity error, on the horizontal slice at the mid-height of a $45\times45\times45$ patch ($180\,\mathrm{m}\times180\,\mathrm{m}$ horizontally). Buildings are shown in gray in the error panels. Right: mean velocity error of 40,000 individually predicted patches, averaged over all 45 layers of each patch, at each horizontal position in the patch. The incoming wind blows along $x$.}}
    \label{fig:patch_examples}
\end{figure}

\begin{figure}[tp]
    \centering
    \includegraphics[width=0.703\linewidth]{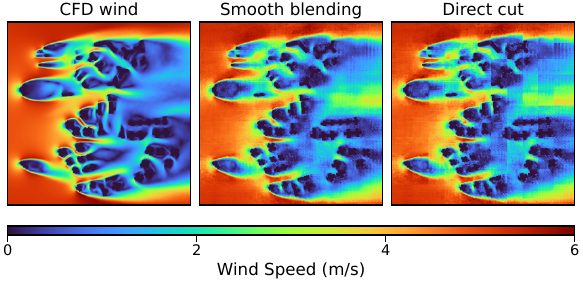}
    \caption{\rev{Horizontal wind speed at $75\,\mathrm{m}$ in urban block A reconstructed from the same patch predictions by smooth blending and by a direct cut at the middle of each overlap, compared with CFD.}}
    \label{fig:stitching}
\end{figure}

\paragraph{Training and inference.}
After computing the full-domain feature fields, we extract local 3D samples of size $45\times45\times45$ grid cells. Each sample contains the geometric features in the patch and uses the co-located CFD velocity patch as the target. Patch locations are randomly sampled during training. The model is trained with an RMSE loss on the predicted velocity field. We use a hidden width of 60 and 12 Fourier modes in each spatial direction. The 3D FNO has four Fourier layers. We train with AdamW, weight decay $10^{-4}$, batch size 16, and an initial learning rate of $10^{-2}$, which is halved every 20 epochs by StepLR. The patch-based training procedure turns each CFD simulation into many local supervision samples while preserving large-scale geometric context through the precomputed SDF and MDDF fields. At test time, the same operator can be evaluated on all patches for full-field prediction or only on corridor-intersecting patches for mission-local planning. Directly stitching non-overlapping patch predictions can introduce boundary discontinuities, which are largely removed by overlapping patches and smooth blending at inference time. The inference time is approximately proportional to the total number of predicted grid cells. All training and testing in this work are conducted on a single $32\,\mathrm{GB}$ NVIDIA V100 GPU. Training takes around 7 hours. Since the available CFD dataset is small, data availability is the main bottleneck. Performance is not highly sensitive to hyperparameter choices as long as severe overfitting is avoided.

\begin{table}[tp]
\revcolor
\centering
\small
\captionsetup{skip=7pt}
\caption{\rev{Mean velocity error $(\mathrm{m/s})$ over all free-space grid points of each evaluation block, for our neural operator with and without MDDF.}}
\label{tab:mddf_ablation}
\begin{tabular}{@{}lccccc@{}}
\toprule
{\bfseries\boldmath Urban block} & {\bfseries\boldmath A} & {\bfseries\boldmath B} & {\bfseries\boldmath C} & {\bfseries\boldmath D} & {\bfseries\boldmath E} \\
\midrule
With MDDF (Ours) & 0.530 & 0.471 & 0.400 & 0.481 & 0.585 \\
Without MDDF & 0.680 & 0.693 & 0.640 & 0.651 & 0.838 \\
\bottomrule
\end{tabular}
\end{table}

\begin{figure}[tp]
    \centering
    \includegraphics[width=0.801\linewidth]{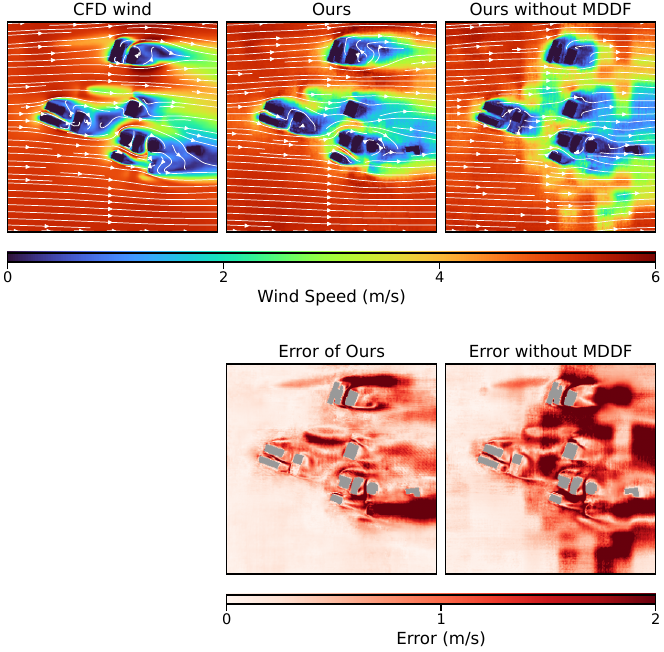}
    \caption{\rev{Horizontal slices at $75\,\mathrm{m}$ in urban block C. Top: horizontal wind speed with streamlines from CFD and from our neural operator with and without MDDF. Bottom: magnitude of the velocity error of the two predictions relative to CFD, with buildings in gray.}}
    \label{fig:mddf_ablation}
\end{figure}

\rev{\paragraph{Patch boundaries and stitching.} A patch predicted on its own is less accurate near its boundary, as Figure~\ref{fig:patch_examples} shows. Measured as the magnitude of the 3D velocity difference from CFD at free-space grid points, the mean error is low in the patch interior and rises toward the boundary, most strongly at the downstream side. At inference time, neighboring patches overlap by at least $40\%$ of the patch size, and each prediction is weighted by a Hann window that decays to nearly zero at the patch boundary. Compared with a direct cut at the middle of each overlap, this smooth blending removes the visible seams (Figure~\ref{fig:stitching}).}

\rev{\paragraph{Effect of MDDF.} We train a second model without the MDDF input channels, with the same architecture and training procedure. Table~\ref{tab:mddf_ablation} compares the mean velocity error of the two models on the five evaluation blocks. Removing MDDF increases the error on every block, by $28\%$ on block A and up to $60\%$ on block C. Figure~\ref{fig:mddf_ablation} shows block C at $75\,\mathrm{m}$. Without MDDF, the predicted wakes behind the buildings are shorter and break up into blocky patches, and the error grows mainly downstream of the buildings.}

\section{Energy consumption model}
\label{sec:energy_model}

This section defines the trajectory energy functional \(E(\tau;\bm{w})\) used
throughout the paper. The same model is used during online planning, where
\(\bm{w}=\bm{w}_p\) is the planning wind field, and during offline evaluation,
where \(\bm{w}=\bm{w}_{\mathrm{CFD}}\) is the high-fidelity CFD wind field. In
GeoWind2Plan, \(\bm{w}_p=\widehat{\bm{w}}_{\mathrm{local}}\); in the wind-agnostic
baseline, \(\bm{w}_p\equiv \bm{0}\). Using the same energy model for all
methods ensures that the reported differences are caused by the planning wind
input rather than by changes in the vehicle model.

We adopt a quasi-steady multirotor power model evaluated along a discretized
trajectory. The model explicitly accounts for spatially varying 3D wind through
the air-relative velocity, and decomposes nodal power into useful mechanical
power, induced power, and blade profile power. The model is intended as a
consistent physics-based planning and evaluation cost; it is not a full
electrochemical battery model and does not model motor-controller transients,
gust loading, or short-time turbulence.

\subsection{Trajectory state and air-relative velocity}

Let a trajectory be discretized into \(N\) nodes. At node \(i\), the UAV state is
\begin{equation}
    \bm{x}_i\in\mathbb{R}^3,\qquad
    \bm{v}_i\in\mathbb{R}^3,\qquad
    \bm{a}_i\in\mathbb{R}^3,
    \label{eq:app_state_variables}
\end{equation}
where \(\bm{x}_i\), \(\bm{v}_i\), and \(\bm{a}_i\) denote position,
ground-relative velocity, and ground-relative acceleration. Given a spatial
wind field \(\bm{w}:\Omega\rightarrow\mathbb{R}^3\), the air-relative velocity
is
\begin{equation}
    \bm{v}_{a,i}
    =
    \bm{v}_i-\bm{w}(\bm{x}_i).
    \label{eq:app_air_relative_velocity}
\end{equation}
This is the quantity through which local wind affects aerodynamic drag, thrust,
and power. Let \(\bm{g}\in\mathbb{R}^3\) denote the gravitational acceleration
vector, pointing downward, with magnitude \(g=\|\bm{g}\|_2\).

\subsection{Aerodynamic drag and required thrust}

We model parasite drag as a quadratic function of air-relative velocity:
\begin{equation}
    \bm{D}_i
    =
    -\frac{1}{2}\rho C_d A_f
    \|\bm{v}_{a,i}\|_2\bm{v}_{a,i},
    \label{eq:app_drag}
\end{equation}
where \(\rho\) is air density, \(C_d\) is an effective drag coefficient, and
\(A_f\) is the frontal reference area. The negative sign indicates that drag
opposes motion relative to the surrounding air.

The thrust vector required to realize the translational acceleration
\(\bm{a}_i\) is obtained from force balance:
\begin{equation}
    \bm{T}_i
    =
    m\bm{a}_i
    -
    m\bm{g}
    -
    \bm{D}_i,
    \label{eq:app_thrust_vector}
\end{equation}
where \(m\) is the UAV mass. We denote the thrust magnitude and thrust-axis unit
vector by
\begin{equation}
    T_i=\|\bm{T}_i\|_2,\qquad
    \revo{\bm{e}_{T,i}}=\frac{\bm{T}_i}{T_i+\revo{\delta}},
    \label{eq:app_thrust_unit}
\end{equation}
where \(\revo{\delta}>0\) is a small numerical constant used only to avoid division
by zero. In normal flight, \(T_i\) is nonzero because the vehicle must at least
counter gravity.

\subsection{Nodal power}

Following standard momentum-theory approximations for rotorcraft, we first
compute the component of the air-relative velocity along the thrust axis:
\begin{equation}
    v_{c,i}
    =
    \bm{v}_{a,i}^{\top}\revo{\bm{e}_{T,i}} .
    \label{eq:app_climb_velocity}
\end{equation}
The induced velocity through the rotor disk is modeled as
\begin{equation}
    v_{\mathrm{ind},i}
    =
    -\frac{1}{2}v_{c,i}
    +
    \sqrt{
        \left(\frac{1}{2}v_{c,i}\right)^2
        +
        \frac{T_i}{2\rho A}
    },
    \label{eq:app_induced_velocity}
\end{equation}
where \(A\) is the total rotor disk area.

The nodal power is decomposed into three components:
\begin{align}
    P_{u,i}
    &=
    \bm{T}_i^{\top}\bm{v}_{a,i},
    \label{eq:app_useful_power}
    \\
    P_{\mathrm{ind},i}
    &=
    T_i v_{\mathrm{ind},i},
    \label{eq:app_induced_power}
    \\
    P_{\mathrm{prof},i}
    &=
    P_{\mathrm{prof},\mathrm{hover}}
    \left(\frac{T_i}{mg}\right)^{3/2}.
    \label{eq:app_profile_power}
\end{align}
Here \(P_{u,i}\) is the useful mechanical power associated with the thrust
force acting through the air-relative velocity. The induced power
\(P_{\mathrm{ind},i}\) accounts for the power required to accelerate air through
the rotor disks, and \(P_{\mathrm{prof},i}\) models blade profile losses scaled
from the hover profile power \(P_{\mathrm{prof},\mathrm{hover}}\).

The total nodal power is
\begin{equation}
    P_i
    =
    P_{u,i}
    +
    P_{\mathrm{ind},i}
    +
    P_{\mathrm{prof},i}.
    \label{eq:app_total_power}
\end{equation}
The useful mechanical term can be negative in certain descending or strongly
wind-assisted configurations, while the induced and profile terms remain
positive. We use the same convention for all planning inputs and all evaluation
conditions. If a deployment requires a \revo{strict} battery-consumption model, a
non-regenerative clamp can be applied to the useful term or to total power, but
that choice should be applied consistently to every method.

\subsection{Trajectory energy}

Given time intervals \(\{\Delta t_i\}_{i=0}^{N-2}\), the trajectory energy under
wind field \(\bm{w}\) is computed by trapezoidal integration:
\begin{equation}
    E(\tau;\bm{w})
    =
    \sum_{i=0}^{N-2}
    \frac{1}{2}
    \left(P_i+P_{i+1}\right)\Delta t_i .
    \label{eq:app_trajectory_energy}
\end{equation}
In the trajectory optimization formulation used in this work, we use a uniform
time step \(\Delta t_i\equiv\Delta t\), which is optimized jointly with the
trajectory states. During CFD evaluation, we evaluate the optimized trajectory
with the same discretization and optimized time step, replacing the planning
wind field by \(\bm{w}_{\mathrm{CFD}}\).

\subsection{Model parameters}

Unless otherwise stated, the parameters in
Eqs.~\eqref{eq:app_drag}--\eqref{eq:app_profile_power} are treated as constants
and set to
\[
    \rho = 1.18\,\mathrm{kg/m^3},\quad
    C_d = 1.0,\quad
    A_f = 0.16\,\mathrm{m^2},\quad
    m = 5.0\,\mathrm{kg},\quad
    A = 0.5\,\mathrm{m^2},
\]
\[
    g = 9.81\,\mathrm{m/s^2},\qquad
    P_{\mathrm{prof},\mathrm{hover}} = 100.0\,\mathrm{W}.
\]
The wind field \(\bm{w}(\bm{x})\) is queried at arbitrary trajectory locations
through interpolation of either the predicted task-local field
\(\widehat{\bm{w}}_{\mathrm{local}}\), the profile baseline, the zero-wind
field, or the CFD field used for offline evaluation.

\section{Energy-aware path planning via RRT initialization and continuous trajectory optimization}
\label{sec:planning_method}

This section provides the full optimization problem used by the planner
\(\mathcal{P}\) in Eq.~\eqref{eq:geometry_wind_planning_map}. Given a planning
wind field \(\bm{w}_p\), a start location \(\bm{x}_s\), a goal location
\(\bm{x}_g\), and the city geometry \(\mathcal{G}\), the planner returns a
collision-free, dynamically feasible trajectory that minimizes the energy
functional \(E(\tau;\bm{w}_p)\). In GeoWind2Plan,
\(\bm{w}_p=\widehat{\bm{w}}_{\mathrm{local}}\); in the CFD-reference planner,
\(\bm{w}_p=\bm{w}_{\mathrm{CFD}}\); in the profile baseline, \(\bm{w}_p\) is
the height-only profile wind; and in the wind-agnostic baseline,
\(\bm{w}_p\equiv\bm{0}\).

The optimization landscape is nonlinear and nonconvex because the power model
depends on local air-relative velocity, the obstacle constraints depend on the
signed distance field, and different feasible corridors may have different wind
exposure. We therefore use a two-stage procedure: RRT first generates diverse
collision-free warm starts, and each warm start is then refined by continuous
trajectory optimization.

\subsection{Environment representation and inputs}

The city geometry is represented by the occupancy field \(\mathcal{G}\) and a
signed distance field \(\mathrm{SDF}_{\mathcal{G}}\). We use the convention
that \(\mathrm{SDF}_{\mathcal{G}}(\bm{x})>0\) in free space,
\(\mathrm{SDF}_{\mathcal{G}}(\bm{x})=0\) on building surfaces, and
\(\mathrm{SDF}_{\mathcal{G}}(\bm{x})<0\) inside buildings. This sign convention
allows clearance constraints to be written as
\(\mathrm{SDF}_{\mathcal{G}}(\bm{x})\ge d_{\mathrm{clr}}\), where
\(d_{\mathrm{clr}}>0\) is the required clearance margin.

The feasible flight workspace is
\begin{equation}
    \Omega_{\mathrm{fly}}
    =
    \left\{
    \bm{x}\in\Omega:
    \mathcal{G}(\bm{x})=0,\;
    z_{\min}\le z \le z_{\max}
    \right\},
    \label{eq:app_flight_workspace_planning}
\end{equation}
with \(z_{\min}\) and \(z_{\max}\) set by the flight envelope. The planner
receives \((\bm{x}_s,\bm{x}_g,\mathcal{G},\mathrm{SDF}_{\mathcal{G}},\bm{w}_p)\)
and returns a trajectory \(\tau\) connecting \(\bm{x}_s\) to \(\bm{x}_g\).

\subsection{Feasible path initialization with RRT}
\label{sec:planning_method_stage1_rrt}

The first stage generates collision-free geometric paths using RRT in the 3D
free space. Starting from \(\bm{x}_s\), the RRT repeatedly samples a point in
\(\Omega_{\mathrm{fly}}\), connects it to the nearest existing tree node by a
short extension, and accepts the edge only if all checked points along the edge
satisfy the clearance constraint. In implementation, the extension length is
adapted using the local SDF value: larger steps are used in open regions, and
smaller steps are used near buildings.

The RRT terminates when it reaches a goal neighborhood around \(\bm{x}_g\),
returning a piecewise-linear waypoint sequence. Because RRT is stochastic, we
run it multiple times to obtain warm starts with different spatial layouts and
homotopy classes. These initial paths are collision-free at the checking
resolution but are not energy-optimal and do not explicitly enforce dynamic
limits. They are used only as feasible initializations for the continuous
optimization stage.

\subsection{Continuous trajectory optimization}
\label{sec:planning_method_stage2}

Each RRT path is resampled into \(N\) trajectory nodes. The optimization
variables are
\begin{equation}
    \mathcal{Z}
    =
    \left\{
    \bm{x}_i,\bm{v}_i,\bm{a}_i
    \right\}_{i=0}^{N-1}
    \cup
    \{\Delta t\},
    \label{eq:app_planning_variables}
\end{equation}
where \(\bm{x}_i\), \(\bm{v}_i\), and \(\bm{a}_i\) are position,
ground-relative velocity, and acceleration, and \(\Delta t\) is a uniform time
step. The planning wind field enters through
\begin{equation}
    \bm{v}_{a,i}
    =
    \bm{v}_i-\bm{w}_p(\bm{x}_i),
    \label{eq:app_planning_air_velocity}
\end{equation}
which determines drag, thrust, and nodal power as defined in
Section~\ref{sec:energy_model}.

\paragraph{Objective.}
For a given planning wind field \(\bm{w}_p\), the continuous refinement solves
\begin{equation}
    \min_{\mathcal{Z}}
    E(\tau(\mathcal{Z});\bm{w}_p)
    =
    \min_{\mathcal{Z}}
    \sum_{i=0}^{N-2}
    \frac{1}{2}
    \left(P_i+P_{i+1}\right)\Delta t,
    \label{eq:app_planning_objective}
\end{equation}
where \(P_i=P(\bm{x}_i,\bm{v}_i,\bm{a}_i,\bm{w}_p(\bm{x}_i))\) is computed
using the energy model in Section~\ref{sec:energy_model}.

\paragraph{Boundary conditions.}
The trajectory endpoints are fixed:
\begin{equation}
    \bm{x}_0=\bm{x}_s,
    \qquad
    \bm{x}_{N-1}=\bm{x}_g .
    \label{eq:app_planning_boundary}
\end{equation}
Endpoint velocities and accelerations are left free but are subject to the same
vehicle feasibility constraints as all other nodes, unless otherwise specified.

\paragraph{Kinematic consistency.}
We enforce trapezoidal collocation:
\begin{align}
    \bm{x}_{i+1}
    &=
    \bm{x}_i
    +
    \frac{1}{2}
    \left(
    \bm{v}_i+\bm{v}_{i+1}
    \right)\Delta t,
    \label{eq:app_planning_kinematics_x}
    \\
    \bm{v}_{i+1}
    &=
    \bm{v}_i
    +
    \frac{1}{2}
    \left(
    \bm{a}_i+\bm{a}_{i+1}
    \right)\Delta t,
    \label{eq:app_planning_kinematics_v}
\end{align}
for \(i=0,\ldots,N-2\).

\paragraph{Obstacle clearance.}
Node-wise clearance is enforced by
\begin{equation}
    \mathrm{SDF}_{\mathcal{G}}(\bm{x}_i)
    \ge
    d_{\mathrm{clr}},
    \qquad
    i=0,\ldots,N-1.
    \label{eq:app_planning_sdf_nodes}
\end{equation}
To avoid cutting through obstacles between adjacent nodes, we also check
intermediate samples on each segment. For
\(\revo{\lambda_k}\in(0,1)\), define
\begin{equation}
    \bm{x}_{i,k}
    =
    (1-\revo{\lambda_k})\bm{x}_i
    +
    \revo{\lambda_k}\bm{x}_{i+1}.
    \label{eq:app_planning_segment_samples}
\end{equation}
We impose
\begin{equation}
    \mathrm{SDF}_{\mathcal{G}}(\bm{x}_{i,k})
    \ge
    d_{\mathrm{clr}},
    \qquad
    i=0,\ldots,N-2,\quad
    k=1,\ldots,K_{\mathrm{seg}}.
    \label{eq:app_planning_sdf_segments}
\end{equation}
\rev{We use $d_{\mathrm{clr}}=10\,\mathrm{m}$ in all experiments. All trajectory nodes and segment samples therefore stay at least $10\,\mathrm{m}$ from building surfaces as a geometric safety margin.}

\paragraph{Workspace and altitude bounds.}
All nodes must remain in the valid map bounds and flight envelope:
\begin{equation}
    \bm{x}_i\in\Omega_{\mathrm{fly}},
    \qquad
    i=0,\ldots,N-1.
    \label{eq:app_planning_workspace}
\end{equation}
In implementation, this is imposed using box bounds on horizontal coordinates
and altitude, together with the SDF clearance constraints above.

\paragraph{Vehicle feasibility.}
The optimized trajectory must satisfy pointwise bounds on airspeed,
acceleration, and thrust:
\begin{equation}
    \|\bm{v}_i-\bm{w}_p(\bm{x}_i)\|_2
    \le
    v_{\max},
    \qquad
    \|\bm{a}_i\|_2
    \le
    a_{\max},
    \qquad
    \|\bm{T}_i\|_2
    \le
    T_{\max},
    \label{eq:app_planning_dynamic_limits}
\end{equation}
for \(i=0,\ldots,N-1\). The thrust \(\bm{T}_i\) is computed from
Eq.~\eqref{eq:app_thrust_vector}. We also bound the time step,
\begin{equation}
    \Delta t_{\min}
    \le
    \Delta t
    \le
    \Delta t_{\max},
    \label{eq:app_planning_time_bounds}
\end{equation}
to avoid degenerate timing solutions and to keep trajectories within the
intended operating regime.

\subsection{Numerical solution and candidate selection}

The resulting problem is a nonlinear program \rev{(NLP)} with nonlinear wind-coupled power
terms and nonlinear SDF constraints. We solve the NLP with an interior-point
optimizer for each RRT warm start. Let
\(\{\tau^{(\revo{\ell})}\}_{\revo{\ell}=1}^{R}\) be the optimized trajectories produced from
\(R\) RRT initializations. The planner selects the candidate with the lowest
planning energy:
\begin{equation}
    \tau^\star
    =
    \arg\min_{\tau^{(\revo{\ell})},\,\revo{\ell}=1,\ldots,R}
    E(\tau^{(\revo{\ell})};\bm{w}_p).
    \label{eq:app_candidate_selection}
\end{equation}

In the experiments, all methods use the same RRT initialization procedure, the
same number of warm starts, the same vehicle limits, the same energy model, and
the same optimizer. Only the planning wind field \(\bm{w}_p\) changes across
methods. After planning, every selected trajectory is evaluated under
\(\bm{w}_{\mathrm{CFD}}\):
\begin{equation}
    E_{\mathrm{CFD}}(\tau^\star)
    =
    E(\tau^\star;\bm{w}_{\mathrm{CFD}}).
    \label{eq:app_cfd_evaluation}
\end{equation}
This protocol directly tests whether the wind information used during planning
leads to lower energy in the high-fidelity CFD wind field.

\subsection{Planning variants}
\label{sec:planning_variants}

All planning variants share the same RRT initialization, continuous trajectory optimizer, vehicle limits, obstacle constraints, and energy model. They differ only in the wind field \(\bm{w}_p\) supplied to the optimizer. GeoWind2Plan uses the task-local predicted 3D wind field, \(\bm{w}_p=\widehat{\bm{w}}_{\mathrm{local}}\). The CFD-reference planner uses the high-fidelity wind field, \(\bm{w}_p=\bm{w}_{\mathrm{CFD}}\), and serves only as a reference because CFD is not available at mission time. The profile-wind baseline uses a height-only field, \(\bm{w}_p=\bm{w}_{\mathrm{profile}}(z)\), which captures the average vertical wind profile but ignores building-induced horizontal variation. The wind-agnostic baseline sets \(\bm{w}_p\equiv\bm{0}\) during optimization.

For fair comparison, every optimized trajectory is evaluated under \(\bm{w}_{\mathrm{CFD}}\). The reported energy therefore measures how useful each planning wind input is for producing a trajectory that performs well in the true building-resolved wind field.

\section{Physical basis of wind-condition adaptation and validation of \revo{inflow-speed} rescaling}
\label{app:speed_generalization}

GeoWind2Plan adapts a neural operator trained under a reference inflow condition to mission-specific incoming background wind vectors. This adaptation has two components with different physical status. Wind-direction adaptation is based on yaw covariance: under yaw-symmetric boundary conditions, rotating the incoming wind direction is equivalent to rotating the city geometry into a reference-wind frame and rotating the predicted vector field back. Wind-speed adaptation is based on an approximate high-Reynolds-number similarity of the time-averaged mean flow, and is therefore validated empirically over the tested speed range. This appendix provides the notation, the yaw-covariance argument, the physical basis for speed rescaling, and the validation metrics used to assess the speed approximation.

\Needspace{8\baselineskip}
\subsection{Notation: reference flow operator}

Let
\[
    \mathcal{S}(\mathcal{G},U\bm{d})
\]
denote the building-resolved, time-averaged urban wind field generated by city
geometry \(\mathcal{G}\) under horizontally uniform inflow direction
\(\bm{d}\in\mathbb{R}^3\), \(\|\bm{d}\|_2=1\), and inflow speed \(U\). Thus
\[
    \mathcal{S}(\mathcal{G},U\bm{d})(\bm{x})
    \in
    \mathbb{R}^3
\]
is the mean wind velocity at location \(\bm{x}\). In the main method,
\(\bm{d}_\theta=(\cos\theta,\sin\theta,0)^\top\) is the mission wind direction,
and \(\bm{e}_1=(1,0,0)^\top\) is the reference inflow direction used by the
neural wind predictor.

\subsection{Yaw covariance of the flow operator}
\label{app:yaw_covariance_proof}

This subsection proves the covariance relation used in
Proposition~\ref{prop:yaw_covariance}. The statement is exact for the idealized
flow operator whose computational domain and boundary conditions are rotated
consistently with the city geometry and inflow. In finite numerical domains, the
same relation is an accurate modeling principle when the outer boundary
conditions are yaw symmetric or sufficiently far from the buildings.

Let \(Q_\theta\in \mathrm{SO}(3)\) be the yaw rotation satisfying
\(Q_\theta\bm{d}_\theta=\bm{e}_1\). For a scalar geometry field
\(\mathcal{G}\), define the rotated geometry
\begin{equation}
    (\mathcal{R}_\theta\mathcal{G})(\bm{y})
    =
    \mathcal{G}(Q_\theta^\top\bm{y}).
    \label{eq:app_rotated_geometry}
\end{equation}

Consider a neutral incompressible steady-flow problem in the free-flow region
outside buildings:
\begin{equation}
    \nabla_{\bm{x}}\cdot\bm{u}=0,
    \qquad
    (\bm{u}\cdot\nabla_{\bm{x}})\bm{u}
    +
    \frac{1}{\rho}\nabla_{\bm{x}}p
    -
    \nu\revo{\nabla^2_{\bm{x}}}\bm{u}
    =
    \bm{0}.
    \label{eq:app_steady_ns}
\end{equation}
The building surfaces satisfy no-slip boundary conditions, and the inflow
boundary imposes a horizontally uniform direction \(U\bm{d}_\theta\), possibly
with a vertical profile that depends on height. The same coordinate-change
argument applies to a unique time- or ensemble-mean turbulent flow field, or to
a yaw-covariant RANS/LES closure, provided the mean solution is unique for the
given forcing.

Let \((\bm{u},p)\) be the solution for geometry \(\mathcal{G}\) and inflow
\(U\bm{d}_\theta\). Define \(\bm{y}=Q_\theta\bm{x}\) and transformed fields
\begin{equation}
    \revo{\bm{u}'}(\bm{y})
    =
    Q_\theta
    \bm{u}(Q_\theta^\top\bm{y}),
    \qquad
    \revo{p'}(\bm{y})
    =
    p(Q_\theta^\top\bm{y}).
    \label{eq:app_rotated_fields}
\end{equation}
Since \(Q_\theta\) is orthogonal, the differential operators transform as
\begin{equation}
    \nabla_{\bm{y}}\revo{\bm{u}'}
    =
    Q_\theta
    (\nabla_{\bm{x}}\bm{u})
    Q_\theta^\top,
    \qquad
    \revo{\nabla^2_{\bm{y}}}\revo{\bm{u}'}
    =
    Q_\theta
    \revo{\nabla^2_{\bm{x}}}\bm{u},
    \qquad
    \nabla_{\bm{y}}\revo{p'}
    =
    Q_\theta
    \nabla_{\bm{x}}p .
    \label{eq:app_derivative_transform}
\end{equation}
Therefore
\(\nabla_{\bm{y}}\cdot\revo{\bm{u}'}=\nabla_{\bm{x}}\cdot\bm{u}=0\), and
\begin{equation}
    (\revo{\bm{u}'}\cdot\nabla_{\bm{y}})\revo{\bm{u}'}
    +
    \frac{1}{\rho}\nabla_{\bm{y}}\revo{p'}
    -
    \nu\revo{\nabla^2_{\bm{y}}}\revo{\bm{u}'}
    =
    Q_\theta
    \left[
    (\bm{u}\cdot\nabla_{\bm{x}})\bm{u}
    +
    \frac{1}{\rho}\nabla_{\bm{x}}p
    -
    \nu\revo{\nabla^2_{\bm{x}}}\bm{u}
    \right]
    =
    \bm{0}.
    \label{eq:app_rotated_ns}
\end{equation}
Thus, the transformed velocity and pressure fields satisfy the same flow
equations in the rotated geometry \(\mathcal{R}_\theta\mathcal{G}\).

The boundary conditions also transform consistently. A no-slip wall remains
no-slip after rotation. The inflow direction becomes
\[
    Q_\theta(U\bm{d}_\theta)
    =
    U\bm{e}_1.
\]
A vertical inflow profile remains a vertical inflow profile because yaw
rotations preserve height. Under yaw-symmetric outer boundary conditions, the
transformed fields are therefore a solution of the reference-direction problem
in the rotated geometry. By uniqueness of the steady solution, or uniqueness of
the time- or ensemble-mean solution,
\begin{equation}
    \revo{\bm{u}'}(\bm{y})
    =
    \mathcal{S}(\mathcal{R}_\theta\mathcal{G},U\bm{e}_1)(\bm{y}).
    \label{eq:app_unique_solution_rotated}
\end{equation}
Substituting \(\bm{y}=Q_\theta\bm{x}\) into
Eq.~\eqref{eq:app_rotated_fields} gives
\begin{equation}
    \mathcal{S}(\mathcal{G},U\bm{d}_\theta)(\bm{x})
    =
    Q_\theta^\top
    \mathcal{S}(\mathcal{R}_\theta\mathcal{G},U\bm{e}_1)
    (Q_\theta\bm{x}).
    \label{eq:app_yaw_covariance_final}
\end{equation}
This proves the yaw-covariance relation used by GeoWind2Plan. The result is a
covariance relation, not an invariance relation: the vector field must be
rotated back to the original coordinate frame.

\subsection{Physical basis of inflow-speed rescaling}
\label{app:physical_basis_speed_rescaling}

Yaw covariance handles wind-direction adaptation. Wind-speed adaptation uses a
different argument: approximate similarity of normalized mean flow in
high-Reynolds-number urban regimes. For readability, we write
\begin{equation}
    \bm{w}(\bm{x};\mathcal{G},U\bm{d})
    =
    \mathcal{S}(\mathcal{G},U\bm{d})(\bm{x}),
    \label{eq:app_w_as_flow_value}
\end{equation}
where \(\mathcal{G}\) is the city geometry, \(\bm{d}\) is a fixed incoming wind
direction, and \(U\) is the inflow speed. For a neutral incompressible flow, nondimensionalizing velocity by \(U\)
and length by a characteristic building height \(H\) introduces the Reynolds
number
\begin{equation}
    \mathrm{Re}
    =
    \frac{UH}{\nu},
    \label{eq:app_reynolds_number}
\end{equation}
where \(\nu\) is the kinematic viscosity of air. In the nondimensional momentum
equation, the viscous term scales as \(1/\mathrm{Re}\). When \(\mathrm{Re}\) is sufficiently
large, the mean flow outside thin near-wall regions is dominated by geometry,
pressure redistribution, separation, and wake interaction rather than the
absolute value of molecular viscosity.

Urban flows around sharp-edged buildings are commonly treated in this
Reynolds-number-independent regime over practical wind speeds. In this regime,
changing \(U\) mainly changes dimensional velocity magnitude, while normalized
mean structures such as channeling, speed-up regions, separation locations, and
recirculation zones remain similar \citep{uehara2003critical,larose2006reynolds,chew2018flows}.
Define the normalized mean field
\begin{equation}
    \widetilde{\bm{w}}(\bm{x};\mathcal{G},U\bm{d})
    =
    \frac{
    \bm{w}(\bm{x};\mathcal{G},U\bm{d})
    }{U}.
    \label{eq:app_normalized_mean_wind}
\end{equation}
The speed-similarity assumption used in the main text is that
\(\widetilde{\bm{w}}\) varies weakly with \(U\) in the operating range tested in
this paper, for fixed \(\mathcal{G}\), fixed \(\bm{d}\), and fixed inflow
profile shape. Therefore,
\begin{equation}
    \bm{w}(\bm{x};\mathcal{G},U\bm{d})
    \approx
    \frac{U}{U_{\mathrm{ref}}}
    \bm{w}(\bm{x};\mathcal{G},U_{\mathrm{ref}}\bm{d}).
    \label{eq:app_speed_rescaling}
\end{equation}

This is not a linearity claim for the Navier--Stokes equations. Instantaneous
flow remains nonlinear, and turbulent fluctuations or thermally driven effects
can depend on Reynolds number and atmospheric conditions. The approximation is
narrower: the normalized, time-averaged mean velocity field used by the current
trajectory optimizer is assumed to be approximately preserved over the tested
wind-speed range.

Combining yaw covariance and speed rescaling yields the wind-condition
adaptation used by \mbox{GeoWind2Plan}:
\begin{equation}
    \bm{w}(\bm{x};\mathcal{G},U\bm{d}_\theta)
    \approx
    \frac{U}{U_{\mathrm{ref}}}
    Q_\theta^\top
    \mathcal{S}
    (\mathcal{R}_\theta\mathcal{G},U_{\mathrm{ref}}\bm{e}_1)
    (Q_\theta\bm{x}).
    \label{eq:app_combined_wind_adaptation}
\end{equation}
The learned neural operator approximates the reference-inflow mapping
\(\mathcal{S}(\cdot,U_{\mathrm{ref}}\bm{e}_1)\), while the rotation and scaling
in Eq.~\eqref{eq:app_combined_wind_adaptation} adapt this reference predictor
to the mission wind vector.

\begin{table}[tp]
    \centering
    \small
    \captionsetup{skip=7pt}
\caption{Metrics for normalized CFD wind-field agreement under speed rescaling. 
    The reference speed is $U_{\mathrm{ref}}=\revo{4\,\mathrm{m/s}}$.}
    \label{tab:speed_rescaling_metrics}
    \begin{tabular}{cccccc}
        \toprule
        Pair & $U_a$ \revo{(m/s)} & $U_b$ \revo{(m/s)} & $U_{\mathrm{ref}}$ \revo{(m/s)} 
        & $\epsilon_{\mathrm{vec}}$ & $\revo{\gamma_{\mathrm{mean}}}$ \\
        \midrule
        $2$ vs. $4$ & $4$ & $2$ & $4$ 
        & $0.0669$ & $0.1115\,\mathrm{rad}$ $(6.39^\circ)$ \\
        $8$ vs. $4$ & $4$ & $8$ & $4$ 
        & $0.1124$ & $0.1365\,\mathrm{rad}$ $(7.82^\circ)$ \\
        \bottomrule
    \end{tabular}
\end{table}

\begin{figure}[tp]
    \centering
    \includegraphics[width=\linewidth]{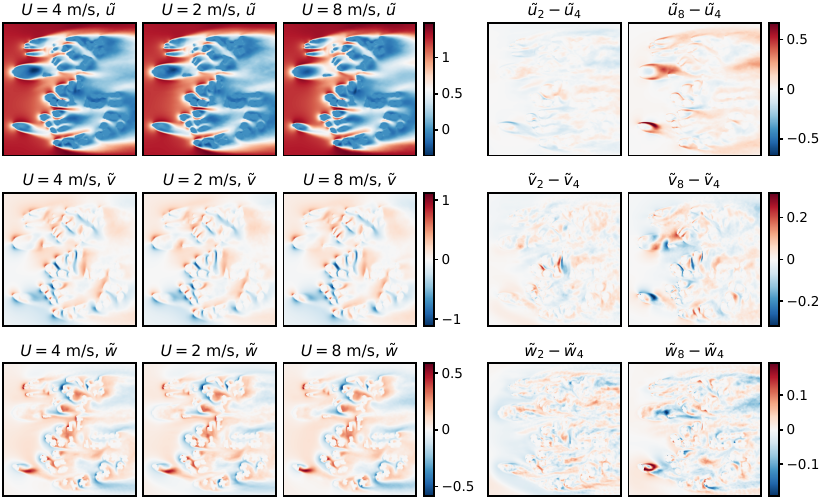}
    \caption{\revo{\textbf{Normalized CFD wind velocity components across varying inlet speeds.}}
    \revo{Normalizing} by the corresponding inlet speeds produces aligned components $(\tilde{u}, \tilde{v}, \tilde{w})$ across different speeds. \rev{The last two columns show the differences of the normalized components between $2$ or $8\,\mathrm{m/s}$ and the reference speed of $4\,\mathrm{m/s}$. The subscript denotes the inlet speed in $\mathrm{m/s}$. The difference columns use narrower color ranges than the field columns. All panels show urban block A at $75\,\mathrm{m}$, and all values are dimensionless.}}
    \label{fig:normalizedwindcomponents}
\end{figure}

\subsection{Metrics and decision-level evaluation for speed rescaling}
\label{app:validation_speed_rescaling}

\rev{We test inflow speeds of $2$, $4$, and $8\,\mathrm{m/s}$, with $4\,\mathrm{m/s}$ as the reference speed. This range covers the most common wind conditions in cities, since long-term observations at the standard measurement height of $10\,\mathrm{m}$, which is also our reference height, show wind speeds mostly below $10\,\mathrm{m/s}$~\cite{he2010probability,zha2017changes}.}
To test the speed-rescaling approximation, we compare CFD mean wind fields
computed at different inflow speeds while holding the city geometry, incoming
wind direction, and boundary-profile shape fixed. Let \(\mathcal{G}\) denote
the fixed city geometry, let \(\bm{d}\) denote the fixed incoming wind
direction, and let \(U_a\) and \(U_b\) be two tested wind speeds. We define the
normalized CFD fields as
\begin{equation}
    \widetilde{\bm{w}}_{\mathrm{CFD}}(\bm{x};\mathcal{G},U_a\bm{d})
    =
    \frac{\bm{w}_{\mathrm{CFD}}(\bm{x};\mathcal{G},U_a\bm{d})}{U_a},
    \qquad
    \widetilde{\bm{w}}_{\mathrm{CFD}}(\bm{x};\mathcal{G},U_b\bm{d})
    =
    \frac{\bm{w}_{\mathrm{CFD}}(\bm{x};\mathcal{G},U_b\bm{d})}{U_b}.
    \label{eq:normalized_cfd_speed_pair}
\end{equation}
The comparison is performed over the free-space UAV flight envelope
\begin{equation}
    \Omega_{\mathrm{fly}}
    =
    \left\{
    \bm{x}\in\Omega:
    \mathcal{G}(\bm{x})=0,\;
    z_{\min}\le z\le z_{\max}
    \right\}.
    \label{eq:app_speed_validation_domain}
\end{equation}

We first compute the relative normalized vector difference
\begin{equation}
    \epsilon_{\mathrm{vec}}
    =
    \frac{
    \left(
    \sum_{\bm{x}\in\Omega_{\mathrm{fly}}}
    \left\|
    \widetilde{\bm{w}}_{\mathrm{CFD}}(\bm{x};\mathcal{G},U_b\bm{d})
    -
    \widetilde{\bm{w}}_{\mathrm{CFD}}(\bm{x};\mathcal{G},U_a\bm{d})
    \right\|_2^2
    \right)^{1/2}
    }{
    \left(
    \sum_{\bm{x}\in\Omega_{\mathrm{fly}}}
    \left\|
    \widetilde{\bm{w}}_{\mathrm{CFD}}(\bm{x};\mathcal{G},U_b\bm{d})
    \right\|_2^2
    \right)^{1/2}
    }.
    \label{eq:speed_rescaling_vector_error}
\end{equation}
This metric evaluates both normalized magnitude and direction differences. We
also compute the mean angular difference
\begin{equation}
    \revo{\gamma_{\mathrm{mean}}}
    =
    \frac{1}{|\Omega_{\mathrm{fly}}|}
    \sum_{\bm{x}\in\Omega_{\mathrm{fly}}}
    \cos^{-1}
    \left(
    \frac{
    \widetilde{\bm{w}}_{\mathrm{CFD}}(\bm{x};\mathcal{G},U_b\bm{d})
    \cdot
    \widetilde{\bm{w}}_{\mathrm{CFD}}(\bm{x};\mathcal{G},U_a\bm{d})
    }{
    \left\|
    \widetilde{\bm{w}}_{\mathrm{CFD}}(\bm{x};\mathcal{G},U_b\bm{d})
    \right\|_2
    \left\|
    \widetilde{\bm{w}}_{\mathrm{CFD}}(\bm{x};\mathcal{G},U_a\bm{d})
    \right\|_2
    +
    \delta
    }
    \right),
    \label{eq:speed_rescaling_angular_error}
\end{equation}
where \(\delta>0\) prevents division by zero near stagnation points. Agreement
between normalized fields supports using the reference-speed prediction after
multiplying by \(U/U_{\mathrm{ref}}\).

In our experiments, the neural operator is trained at the reference speed
\(U_{\mathrm{ref}}\). Let
\(\widehat{\bm{w}}_{\mathrm{local}}^{\mathrm{ref}}(\bm{x};\bm{d}_{\theta})\)
denote the task-local field after rotating the city into the reference-wind
frame, applying the reference-speed neural operator, stitching the selected
patches, and rotating the vector field back, but before speed rescaling. For a
mission background speed \(U\), the planning wind field is
\begin{equation}
    \widehat{\bm{w}}_{\mathrm{local}}(\bm{x};U,\bm{d}_{\theta})
    =
    \frac{U}{U_{\mathrm{ref}}}
    \widehat{\bm{w}}_{\mathrm{local}}^{\mathrm{ref}}(\bm{x};\bm{d}_{\theta}).
    \label{eq:app_predicted_local_speed_rescaling}
\end{equation}
The main experiments further validate this approximation at the decision level
by planning with the scaled predicted wind field and evaluating the resulting
trajectory under CFD wind at the corresponding mission background speed.

Decision-level results for \(2\), \(4\), and \(8\,\mathrm{m/s}\) are reported in \revo{Section}~\ref{sec:energy_performance}\revo{.}

Table~\ref{tab:speed_rescaling_metrics} shows that, after normalizing the CFD wind fields by the corresponding inflow speed, the fields at different speeds remain relatively close to the reference-speed field. For the comparison between $U=2\,\mathrm{m/s}$ and $U_{\mathrm{ref}}=4\,\mathrm{m/s}$, the relative normalized vector difference is only $6.69\%$, with a mean angular difference of $6.39^\circ$. For the higher-speed case $U=8\,\mathrm{m/s}$, the error increases to $11.24\%$, but the mean angular difference remains below $8^\circ$. These results indicate that the normalized wind fields are not identical, but their magnitudes and directions are sufficiently consistent to support the speed-rescaling approximation used in Eq.~(\ref{eq:app_predicted_local_speed_rescaling}). The larger discrepancy at $U=8\,\mathrm{m/s}$ suggests that nonlinear flow effects become more noticeable as the mission background speed moves farther from the reference speed, so the approximation should still be checked at the decision level through trajectory evaluation under the corresponding CFD wind field.

\revo{Figure~\ref{fig:normalizedwindcomponents} provides} visual evidence for the validity of the speed-rescaling approximation.
After each wind field is divided by its corresponding inlet speed, Figure~\ref{fig:normalizedwindcomponents} shows that the normalized velocity components $(\tilde{u},\tilde{v},\tilde{w})$ align closely across all tested speeds.
This demonstrates that changes in inlet speed primarily rescale the wind-field magnitude while preserving the underlying spatial flow structure. \rev{The differences between the normalized components, shown in the last two columns of Figure~\ref{fig:normalizedwindcomponents}, are small over most of the domain, are concentrated in building wakes, and are larger at $8\,\mathrm{m/s}$ than at $2\,\mathrm{m/s}$, consistent with Table~\ref{tab:speed_rescaling_metrics}.}
Together with the quantitative results in Table~\ref{tab:speed_rescaling_metrics}, these visual comparisons strongly support the use of the reference-speed wind prediction scaled by $U/U_{\mathrm{ref}}$ for planning at different background wind speeds.

\subsection{Scope and limitations}
\label{app:speed_rescaling_scope}

The yaw-covariance relation supports adaptation across horizontal incoming wind directions when the geometry, inflow, and outer boundary conditions are rotated consistently. In a finite CFD domain, this relation can be affected by non-yaw-symmetric boundary treatments, finite-domain blockage, interpolation errors introduced by rotating voxelized geometry, or terrain and boundary conditions that are not horizontally homogeneous. Thus, the direction transformation should be interpreted as a physically structured adaptation principle, not as a guarantee for every possible numerical setup.

The speed-rescaling validation supports speed-magnitude adaptation for the time-averaged mean wind field under fixed inflow direction, fixed boundary-profile shape, and neutral mechanically driven flow. It should not be interpreted as validation for thermally stratified atmospheric boundary layers, gusty inflow, unsteady turbulence statistics, or arbitrary sensor noise in the background wind estimate. These effects require additional model inputs, additional CFD data, or separate validation.

Within these assumptions, the wind-condition adaptation used by GeoWind2Plan is aligned with the needs of the trajectory optimizer, which uses the time-averaged mean wind field through the air-relative velocity rather than directly modeling short-time turbulent fluctuations.

\section{Compute resources}
\label{app:compute}
All neural-operator experiments were run on a single NVIDIA V100 GPU with $32\,\mathrm{GB}$ memory. Training the final model required approximately 7 GPU-hours. For inference on a $1.2\,\mathrm{km}$ urban block, full-field neural prediction took about $15\,\mathrm{s}$, while 20\% corridor-localized prediction took about $3\,\mathrm{s}$. \rev{These $3\,\mathrm{s}$ include corridor extraction, neural operator inference, patch stitching, and the rotation and scaling of the wind field. The geometric input features are computed before the mission and are reused for all start-goal pairs.} The CFD/LES labels were generated using CityFFD on the same hardware, i.e., a single NVIDIA V100 GPU with $32\,\mathrm{GB}$ memory. Generating one high-resolution $1.2\,\mathrm{km}$ urban-block CFD/LES case took approximately 8 GPU-hours. \rev{Each CFD field is valid for only one urban block and one inflow condition, and a new wind direction or speed requires a new simulation. GeoWind2Plan uses one trained model for different wind directions and speeds.} The main storage cost comes from saving large LES trajectories. For example, storing 500 snapshots for one $1.2\,\mathrm{km}$ urban-block LES simulation requires approximately $600\,\mathrm{GB}$. 

In contrast, the trajectory-planning component is computationally lightweight and was run on CPUs. For each planning instance, one RRT initialization followed by trajectory optimization took approximately $2\,\mathrm{s}$ on a CPU core. Since the 10 RRT initializations used for each mission are independent, they can be parallelized over 10 CPU cores, giving an approximate wall-clock planning time of $2\,\mathrm{s}$ per mission, or about 20 CPU-seconds if executed serially. Overall, the computational cost of the reported pipeline is dominated by CFD/LES data generation and neural operator training, while trajectory optimization is relatively small.

\section{Notation}
\label{app:notation}

This section summarizes the main notation used in the method and appendices.
Scalars are written in italic letters, vectors in bold letters, sets and
function spaces in calligraphic letters, and learned or physical operators in
capital letters.

\renewcommand{\arraystretch}{1.12}
\begin{longtable}{@{}p{0.24\linewidth}p{0.73\linewidth}@{}}
\caption{\rev{Notation used in the paper.}}\label{tab:notation}\\
\toprule
\textbf{Symbol} & \textbf{Meaning} \\
\midrule
\endfirsthead

\caption*{\rev{Table~\thetable: Notation used in the paper (continued).}}\\
\toprule
\textbf{Symbol} & \textbf{Meaning} \\
\midrule
\endhead

\midrule
\multicolumn{2}{@{}r@{}}{\rev{\footnotesize\textit{Continued on next page}}} \\
\endfoot

\bottomrule
\endlastfoot

\multicolumn{2}{@{}l}{\textit{Urban domain and geometry}} \\*
\(\Omega\subset\mathbb{R}^3\) 
& 3D urban computational domain. \\
\(\Omega_{\mathrm{fly}}\) 
& Feasible UAV flight envelope, restricted to free space and the allowed altitude band. \\
\(\Omega_{\mathrm{task}}\) 
& Task-relevant region used for local wind inference and planning. \\
\leavevmode\rev{\(d_{sg}\), \(r\)} 
& \leavevmode\rev{Horizontal start-goal distance and relative corridor width, giving a total corridor width of \(r d_{sg}\).} \\
\(\bm{x}\), \(\bm{y}\) 
& Spatial locations in the original city frame and the reference-wind frame, respectively. \\
\(\mathcal{G}\) 
& Building occupancy field; \(\mathcal{G}(\bm{x})=1\) denotes building space and \(\mathcal{G}(\bm{x})=0\) denotes free space. \\
\(\mathrm{SDF}_{\mathcal{G}}\) 
& Signed distance field of the city geometry, positive in free space under our convention. \\
\(\mathrm{MDDF}_{\mathcal{G}}^{(k)}\) 
& The \(k\)-th multi-directional distance feature, measuring obstacle distance along a prescribed direction. \\
\(\mathcal{H}(\mathcal{G})\) 
& Building-aware geometry feature field, consisting of occupancy, SDF, and MDDF features. \\

\midrule
\multicolumn{2}{@{}l}{\textit{Background wind and coordinate transformation}} \\*
\(\bm{u}_{\mathrm{bg}}\) 
& Incoming background wind vector available at mission time,
\(\bm{u}_{\mathrm{bg}}=U\bm{d}_{\theta}\). \\
\(U\) 
& Incoming background wind speed magnitude. \\
\(\theta\) 
& Incoming background wind direction in the horizontal plane. \\
\(\bm{d}_{\theta}\) 
& Unit vector of the incoming background wind direction,
\(\bm{d}_{\theta}=(\cos\theta,\sin\theta,0)^\top\). \\
\(\psi\) 
& Relative mission-wind angle between the horizontal \revo{start-goal} direction and the background wind direction; used for tailwind/headwind/crosswind grouping. \\
\(\bm{e}_1\) 
& Reference inflow direction, \(\bm{e}_1=(1,0,0)^\top\). \\
\(U_{\mathrm{ref}}\) 
& Reference inflow speed used to train the neural wind predictor. \\
\(Q_{\theta}\in \mathrm{SO}(3)\) 
& Yaw rotation that maps the mission wind direction to the reference direction, \(Q_{\theta}\bm{d}_{\theta}=\bm{e}_1\). \\
\(\mathcal{R}_{\theta}\mathcal{G}\) 
& Rotated city geometry in the reference-wind frame,
\((\mathcal{R}_{\theta}\mathcal{G})(\bm{y})=\mathcal{G}(Q_{\theta}^{\top}\bm{y})\). \\

\midrule
\multicolumn{2}{@{}l}{\textit{Wind prediction}} \\*
\(\mathcal{S}(\mathcal{G},U\bm{d})\) 
& Ideal building-resolved mean-flow solution operator for geometry \(\mathcal{G}\), incoming wind speed \(U\), and incoming direction \(\bm{d}\). Its value \(\mathcal{S}(\mathcal{G},U\bm{d})(\bm{x})\in\mathbb{R}^3\) is the time-averaged wind velocity at location \(\bm{x}\). \\

\(\bm{w}(\bm{x};\mathcal{G},U\bm{d})\) 
& Generic time-averaged urban wind field at location \(\bm{x}\) for geometry \(\mathcal{G}\), incoming wind speed \(U\), and incoming direction \(\bm{d}\). In Appendix~\ref{app:speed_generalization}, \(\bm{w}(\bm{x};\mathcal{G},U\bm{d})=\mathcal{S}(\mathcal{G},U\bm{d})(\bm{x})\). \\

\(\bm{w}_{\mathrm{CFD}}(\bm{x};\mathcal{G},\bm{u}_{\mathrm{bg}})\) 
& High-fidelity CFD wind field for geometry \(\mathcal{G}\) and incoming background wind vector \(\bm{u}_{\mathrm{bg}}\), used for offline evaluation and CFD-reference planning. Equivalently, when \(\bm{u}_{\mathrm{bg}}=U\bm{d}\), this field can be written as \(\bm{w}_{\mathrm{CFD}}(\bm{x};\mathcal{G},U\bm{d})\). \\

\(\widehat{\bm{w}}_{\mathrm{local}}\) 
& Task-local predicted 3D wind field produced by GeoWind2Plan. \\

\(\widehat{\bm{w}}_{\mathrm{local}}^{\mathrm{ref}}\) 
& Task-local predicted wind field after wind-direction adaptation and vector rotation back to the city frame, but before speed rescaling. \\

\(\bm{w}_p\) 
& Planning wind field supplied to the trajectory optimizer. \\
\(\bm{w}_{\mathrm{profile}}(z)\) 
& Height-only wind-profile baseline with no horizontal spatial structure. \\
\(F_{\phi}\) 
& Localized geometry-conditioned neural operator trained under the reference inflow condition. \\
\(\Pi_{\phi}\) 
& Full task-local wind prediction module, including geometry transformation, patch inference, stitching, and speed scaling. \\
\(\revo{\Omega_j}\), \(\mathcal{J}_{\mathrm{task}}\) 
& Local 3D patch \(\revo{\Omega_j}\) and the set of task-relevant patch indices evaluated at mission time. \\
\(\widetilde{\bm{w}}\) 
& Normalized mean wind field, typically
\(\widetilde{\bm{w}}(\bm{x};\mathcal{G},U\bm{d})
=
\bm{w}(\bm{x};\mathcal{G},U\bm{d})/U\). \\

\midrule
\multicolumn{2}{@{}l}{\textit{Trajectory planning}} \\*
\(\bm{x}_s,\bm{x}_g\) 
& Start and goal locations. \\
\(\tau\) 
& UAV trajectory,
\(\tau=\{\bm{x}(t),\bm{v}(t)\}_{t\in[0,\revo{t_{\mathrm f}}]}\). \\
\(\revo{t_{\mathrm f}}\) 
& Final mission time. \\
\(\mathcal{T}(\mathcal{G},\bm{x}_s,\bm{x}_g)\) 
& Set of collision-free and dynamically feasible trajectories connecting \(\bm{x}_s\) and \(\bm{x}_g\). \\
\(\mathcal{P}\) 
& Trajectory planner. \\
\(\tau^\star\) 
& Selected optimized trajectory returned by the planner. \\
\(N\) 
& Number of trajectory discretization nodes. \\
\(\bm{x}_i,\bm{v}_i,\bm{a}_i\) 
& Position, ground-relative velocity, and acceleration at node \(i\). \\
\(\Delta t\) 
& Uniform time step optimized jointly with the trajectory states. \\
\(\mathcal{Z}\) 
& Trajectory optimization variables,
\(\mathcal{Z}=\{\bm{x}_i,\bm{v}_i,\bm{a}_i\}_{i=0}^{N-1}\cup\{\Delta t\}\). \\
\(d_{\mathrm{clr}}\) 
& Required obstacle-clearance margin. \\
\(v_{\max},a_{\max},T_{\max}\) 
& Bounds on airspeed, acceleration magnitude, and thrust magnitude. \\

\pagebreak
\multicolumn{2}{@{}l}{\textit{Energy model}} \\*
\(E(\tau;\bm{w})\) 
& Physical energy consumed by trajectory \(\tau\) under wind field \(\bm{w}\). \\
\(E_{\mathrm{CFD}}(\tau)\) 
& CFD-evaluated trajectory energy,
\(E_{\mathrm{CFD}}(\tau)=E(\tau;\bm{w}_{\mathrm{CFD}})\). \\
\(\bm{v}_{a,i}\) 
& Air-relative velocity at node \(i\),
\(\bm{v}_{a,i}=\bm{v}_i-\bm{w}(\bm{x}_i)\). \\
\leavevmode\rev{\(\rho\), \(C_d\), \(A_f\)} 
& \leavevmode\rev{Air density, effective drag coefficient, and frontal reference area.} \\
\leavevmode\rev{\(m\), \(\bm{g}\)} 
& \leavevmode\rev{UAV mass and gravitational acceleration vector.} \\
\(\bm{D}_i\) 
& Parasite drag force at node \(i\). \\
\(\bm{T}_i\), \(T_i\) 
& Required thrust vector and its magnitude at node \(i\). \\
\(P_i\) 
& Total nodal power at node \(i\). \\
\(P_{u,i},P_{\mathrm{ind},i},P_{\mathrm{prof},i}\) 
& Useful mechanical power, induced power, and blade profile power at node \(i\). \\

\midrule
\multicolumn{2}{@{}l}{\textit{Evaluation}} \\*
\(\mathrm{Re}\) 
& Reynolds number used in the speed-rescaling discussion,
\(\mathrm{Re}=UH/\nu\). \\
\(\epsilon_{\mathrm{vec}}\), \(\revo{\gamma_{\mathrm{mean}}}\) 
& Relative normalized vector difference and mean angular difference used to compare normalized CFD wind fields at different incoming wind speeds. \\
\(\bm{0}\) 
& Zero wind field used by the wind-agnostic baseline. \\
\(H\) 
& Characteristic building height used in the Reynolds number definition for the speed-rescaling discussion. \\

\(\nu\) 
& Kinematic viscosity of air used in the Reynolds number definition. \\

\(\delta\) 
& Small positive constant used to avoid division by zero when computing angular differences near stagnation points. \\

\end{longtable}

\newpage
\section*{NeurIPS Paper Checklist}

\begin{enumerate}

\item {\bf Claims}
    \item[] Question: Do the main claims made in the abstract and introduction accurately reflect the paper's contributions and scope?
    \item[] Answer: \answerYes{}
    \item[] Justification: The abstract and introduction clearly state the framework’s scope, inputs, outputs, main contributions, and evaluation protocol, including the emphasis on decision-useful wind prediction rather than CFD-perfect reconstruction. The experimental claims about inference time and energy savings are reported consistently in the results. \rev{See the abstract, Section~\ref{sec:intro}, and Section~\ref{sec:energy_performance}.}

\item {\bf Limitations}
    \item[] Question: Does the paper discuss the limitations of the work performed by the authors?
    \item[] Answer: \answerYes{}
    \item[] Justification: The paper includes an explicit limitations discussion, noting that it uses time-averaged wind fields, does not model turbulence-induced power fluctuations or safety risks, and assumes fixed background wind during a mission. \rev{See Appendix~\ref{app:limitations}.}

\item {\bf Theory assumptions and proofs}
    \item[] Question: For each theoretical result, does the paper provide the full set of assumptions and a complete (and correct) proof?
    \item[] Answer: \answerYes{}
    \item[] Justification: The main theoretical result, the yaw-covariance proposition, states its assumptions and is proved in Appendix~\ref{app:yaw_covariance_proof}. The wind-speed rescaling component is correctly framed as an approximation/assumption rather than a theorem, with physical justification and empirical validation provided separately.

    \item {\bf Experimental result reproducibility}
    \item[] Question: Does the paper fully disclose all the information needed to reproduce the main experimental results of the paper to the extent that it affects the main claims and/or conclusions of the paper (regardless of whether the code and data are provided or not)?
    \item[] Answer: \answerYes{}
    \item[] Justification: The paper provides the main ingredients needed to reproduce the experiments, including the model design, CFD data-generation procedure, training hyperparameters, energy model, planning variants, baselines, and CFD-based evaluation protocol. \rev{See Section~\ref{sec:methodology} and Appendices~\ref{sec:model_detail}, \ref{sec:energy_model}, and~\ref{sec:planning_method}.}

\item {\bf Open access to data and code}
    \item[] Question: Does the paper provide open access to the data and code, with sufficient instructions to faithfully reproduce the main experimental results, as described in supplemental material?
    \item[] Answer: \answerYes{}
    \item[] Justification: The code and data are publicly available at \url{https://github.com/DUAL-Xiao/GeoWind2Plan/} and on Hugging Face (Appendix~\ref{sec:model_detail}), and the appendix gives the details needed to reproduce the main experiments.

\item {\bf Experimental setting/details}
    \item[] Question: Does the paper specify all the training and test details (e.g., data splits, hyperparameters, how they were chosen, type of optimizer) necessary to understand the results?
    \item[] Answer: \answerYes{}
    \item[] Justification: The paper specifies the evaluated urban blocks, wind speeds, mission sampling, flight envelope, baselines, RRT initializations, model training details, optimizer structure, and evaluation protocol. \rev{See Section~\ref{sec:experimental_setup} and Appendix~\ref{sec:model_detail}.}

\item {\bf Experiment statistical significance}
    \item[] Question: Does the paper report error bars suitably and correctly defined or other appropriate information about the statistical significance of the experiments?
    \item[] Answer: \answerYes{}
    \item[] Justification: The main quantitative tables report mean and standard deviation over the evaluated missions and relative overheads against the CFD-reference planner. \rev{See Table~\ref{tab:large_experiment} and Appendix Table~\ref{tab:large_experiment_supp}.}

\item {\bf Experiments compute resources}
    \item[] Question: For each experiment, does the paper provide sufficient information on the computer resources (type of compute workers, memory, time of execution) needed to reproduce the experiments?
    \item[] Answer: \answerYes{}
    \item[] Justification: The paper provides sufficient compute-resource information to reproduce the experiments, including the worker type, GPU memory, training and inference time, CFD/LES generation time, storage cost, and CPU planning time. \rev{See Appendix~\ref{app:compute}.}

\item {\bf Code of ethics}
    \item[] Question: Does the research conducted in the paper conform, in every respect, with the NeurIPS Code of Ethics \url{https://neurips.cc/public/EthicsGuidelines}?
    \item[] Answer: \answerYes{}
    \item[] Justification: The work uses simulated/CFD-generated urban wind fields and UAV planning experiments, does not involve human subjects or personal data, and does not appear to violate the NeurIPS Code of Ethics. \rev{See Appendix~\ref{sec:model_detail} for the data.}

\item {\bf Broader impacts}
    \item[] Question: Does the paper discuss both potential positive societal impacts and negative societal impacts of the work performed?
    \item[] Answer: \answerYes{}
    \item[] Justification: The paper includes a broader impacts discussion covering both potential benefits for energy-efficient urban UAV operations and potential risks from larger-scale drone deployment, including noise, low-altitude airspace congestion, and privacy concerns. \rev{See Appendix~\ref{app:broader_impacts}.}

\item {\bf Safeguards}
    \item[] Question: Does the paper describe safeguards that have been put in place for responsible release of data or models that have a high risk for misuse (e.g., pre-trained language models, image generators, or scraped datasets)?
    \item[] Answer: \answerNA{}
    \item[] Justification: The paper does not appear to release high-risk assets such as pretrained language models, image generators, or scraped datasets requiring special safeguards.

\item {\bf Licenses for existing assets}
    \item[] Question: Are the creators or original owners of assets (e.g., code, data, models), used in the paper, properly credited and are the license and terms of use explicitly mentioned and properly respected?
    \item[] Answer: \answerYes{}
    \item[] Justification: The paper cites the existing models, datasets, and software packages used in the experiments. \rev{See Appendices~\ref{sec:model_detail} and~\ref{app:additional_baselines}.}

\item {\bf New assets}
    \item[] Question: Are new assets introduced in the paper well documented and is the documentation provided alongside the assets?
    \item[] Answer: \answerYes{}
    \item[] Justification: The paper releases the GeoWind2Plan code and the CFD dataset, available at \url{https://github.com/DUAL-Xiao/GeoWind2Plan/} and on Hugging Face (Appendix~\ref{sec:model_detail}).

\item {\bf Crowdsourcing and research with human subjects}
    \item[] Question: For crowdsourcing experiments and research with human subjects, does the paper include the full text of instructions given to participants and screenshots, if applicable, as well as details about compensation (if any)? 
    \item[] Answer: \answerNA{}
    \item[] Justification: The paper does not involve crowdsourcing experiments or research with human subjects.

\Needspace{6\baselineskip}
\item {\bf Institutional review board (IRB) approvals or equivalent for research with human subjects}
    \item[] Question: Does the paper describe potential risks incurred by study participants, whether such risks were disclosed to the subjects, and whether Institutional Review Board (IRB) approvals (or an equivalent approval/review based on the requirements of your country or institution) were obtained?
    \item[] Answer: \answerNA{}
    \item[] Justification: The paper does not involve human-subject research, so IRB or equivalent approval is not applicable.

\item {\bf Declaration of LLM usage}
    \item[] Question: Does the paper describe the usage of LLMs if it is an important, original, or non-standard component of the core methods in this research? Note that if the LLM is used only for writing, editing, or formatting purposes and does \emph{not} impact the core methodology, scientific rigor, or originality of the research, declaration is not required.
    \item[] Answer: \answerNA{}
    \item[] Justification: The core method does not use LLMs as an important, original, or non-standard component.

\end{enumerate}

\end{document}